\documentclass[journal]{IEEEtran}

\ifCLASSINFOpdf
   \usepackage[pdftex]{graphicx}
   \DeclareGraphicsExtensions{.pdf,.jpeg,.png}
\else
\fi

\usepackage{amsmath}
\usepackage{amssymb}

\usepackage{tikz}
\usetikzlibrary{matrix,decorations.pathreplacing, calc, positioning}
\usepackage{multirow}
\usepackage{bm}
\usepackage{url}
\usepackage{float}
\usepackage{diagbox}

\DeclareMathOperator*{\argmin}{argmin}
\DeclareMathOperator*{\argmax}{argmax}
\newcommand*{\argminl}{\argmin\limits}
\newcommand*{\argmaxl}{\argmax\limits}

\newcommand{\del}[1]{}

\renewcommand{\tabcolsep}{2pt}

\usepackage[linesnumbered, ruled, vlined]{algorithm2e}

\ifCLASSOPTIONcompsoc
 \usepackage[caption=false,font=normalsize,labelfont=sf,textfont=sf]{subfig}
\else
 \usepackage[caption=false,font=footnotesize]{subfig}
\fi

\begin{document}

\title{Adaptive Kernel Value Caching for SVM Training}

\author{Qinbin Li,
        Zeyi Wen$^*$,
        Bingsheng He$^*$
\thanks{Q. Li and B. He are with National University of Singapore. Email: \{qinbin, hebs\}@comp.nus.edu.sg}%
\thanks{Z. Wen is with The Univerisity of Western Australia. Email: zeyi.wen@uwa.edu.au}%
\thanks{Z. Wen and B. He are the corresponding authors.}%
\thanks{Digital Object Identifier 10.1109/TNNLS.2019.2944562~\copyright~2019 IEEE}
}

\markboth{}%
{Shell \MakeLowercase{\textit{et al.}}: Bare Demo of IEEEtran.cls for IEEE Journals}

\maketitle

\begin{abstract}
Support Vector Machines (SVMs) can solve structured multi-output learning problems such as multi-label classification, multiclass classification and vector regression. SVM training is expensive especially for large and high dimensional datasets. The bottleneck of the SVM training often lies in the kernel value computation. In many real-world problems, the same kernel values are used in many iterations during the training, which makes the caching of kernel values potentially useful. The majority of the existing studies simply adopt the LRU (\underline{l}east \underline{r}ecently \underline{u}sed) replacement strategy for caching kernel values. However, as we analyze in this paper, the LRU strategy generally achieves high hit ratio near the final stage of the training, but does not work well in the whole training process. Therefore, we propose a new caching strategy called EFU (l\underline{e}ss \underline{f}requently \underline{u}sed) which replaces the less frequently used kernel values that enhances LFU (\underline{l}east \underline{f}requently \underline{u}sed). Our experimental results show that EFU often has 20\% higher hit ratio than LRU in the training with the Gaussian kernel. To further optimize the strategy, we propose a caching strategy called HCST (\underline{h}ybrid \underline{c}aching for the \underline{S}VM \underline{t}raining), which has a novel mechanism to automatically adapt the better caching strategy in the different stages of the training. We have integrated the caching strategy into ThunderSVM, a recent SVM library on many-core processors. Our experiments show that HCST adaptively achieves high hit ratios with little runtime overhead among different problems including multi-label classification, multiclass classification and regression problems. Compared with other existing caching strategies, HCST achieves 20\% more reduction in training time on average.
\end{abstract}


\IEEEpeerreviewmaketitle

\section{Introduction}
The Support Vector Machine (SVM)~\cite{cortes1995support} is a classic supervised machine learning algorithm, and can solve problems with structured, unstructured and semi-structured data~\cite{kashima2002kernels}. SVMs can solve structured multi-output learning problems, which include multi-label classification~\cite{wu2015ml,wu2016ml,wen2018efficient}, multiclass classification~\cite{li2013active, liu2015optimality} and vector regression~\cite{xu2013multi}. Some examples applications of SVMs include document classification, object detection, and image classification~\cite{liu2018hyperspectral}. The underlying idea of training SVMs is to find a hyperplane to separate the two classes of data in their original data space. To handle non-linearly separable data, SVMs use a kernel function~\cite{scholkopf2001learning} to map data to a higher dimensional space, where the data may become linearly separable.

Although SVMs have several intriguing properties, the high training cost for large datasets is a deficiency. Even though many existing studies have been done for accelerating the SVM training~\cite{platt1998sequential,tsang2005core,wenthundersvm18}, the kernel value computation is usually the most time-consuming operation of the training. To reduce the cost of kernel value computation, caching kernel values may be a good solution. Since the same kernel values are often used in different iterations during the training, we can avoid computing the kernel values if they are cached. Many SVM libraries such as LIBSVM and SVM$^{light}$~\cite{joachims1999svmlight} adopt the LRU replacement strategy for caching kernel values. LRU works well for occasions with good temporal locality, which may not happen in the SVM training. In order to analyze the patterns in the entire SVM training, we divide the training process into several stages evenly according to the number of iterations. As we observed in the experiments, LRU only works well near the final stage of the training. 

However, proposing a suitable caching strategy for the SVM training is challenging, because (i) the access pattern is not trivial to identify, and (ii) the caching strategy should be lightweight in terms of runtime overhead. Based on the pattern analysis, we propose a new strategy EFU that enhances LFU. The EFU strategy replaces the l\underline{e}ss \underline{f}requently \underline{u}sed kernel values when the cache is full, which is suitable for the access pattern of the kernel values. In order to adapt the access pattern of different stages, we propose the HCST replacement strategy, which automatically switches to the better strategy between EFU and LRU using the collected statistics. We collect the exact number of cache hits of the strategy being used and estimate the approximate number of cache hits of the other strategy based on its characteristic. Thus, HCST can automatically switch the strategy between EFU and LRU. To reduce the cost of copying data to cache, we perform the replacement of HCST in parallel. Compared with the existing strategies, the HCST strategy can achieve 20\% more reduction in training time on average.

The main contributions of this paper are as follows.

\begin{itemize}

\item By splitting the training process to stages, we discover common features for the access patterns of different datasets.

\item We propose a new caching strategy, EFU, which enhances LFU to take advantage of the access patterns of the kernel values.

\item We design an adaptive caching scheme, HCST, which can fully utilize the characteristics of EFU and LRU to achieve a better performance.

\item We conduct experiments on different problems including multi-label classification, multiclass classification and regression problems. The experimental results show that HCST is superior compared with other caching strategies, including LRU, LFU, LAT~\cite{wen2014mascot} and EFU.
\end{itemize}

\section{Preliminaries and related work}
\label{paper:preli}

In this section, we first present the formal definition of the SVM training problem, and explain a commonly used SVM training algorithm called \textit{SMO}. Then, we describe a more recent SVM training algorithm~\cite{wenthundersvm18} which solves multiple SMO subproblems in each iteration and exploits batch processing. Finally, we discuss the existing SVM libraries and caching strategies.

\subsection{The SVM training problem}
A training instance $x_i$ is attached with an integer $y_i\ \in\ \{+1, -1\}$ as its label. A positive (negative) instance is an instance with the label of $+1$ ($-1$). Given a set $\boldsymbol{\chi}$ of $n$ training instances, the goal of the SVM training is to find a hyperplane that separates the positive and the negative training instances in the feature space induced by the kernel function with the maximum margin and meanwhile, with the minimum misclassification error on the training instances. The SVM training is equivalent to solving the following problem:

\vspace{-7pt}
\begin{equation}
\begin{aligned}
&\underset{\boldsymbol{\alpha}}{\text{max}}
& & \sum_{i=1}^{n}{\alpha_i}-\frac{1}{2}{\boldsymbol{\alpha^T} \boldsymbol{Q} \boldsymbol{\alpha}} \\
& \text{subject to}
& &  0 \leq \alpha_i \leq C, \forall i \in \{1,...,n\}, \sum_{i=1}^{n}{y_i\alpha_i} = 0
\end{aligned}
\label{eq:svm_dual}
\end{equation}
\vspace{-7pt}

where {$\boldsymbol{\alpha} \in \mathbb{R}^n$} is a weight vector, 
and $\alpha_i$ denotes the \textit{weight} of $\boldsymbol{x}_i$; {$C$} is for regularization;
{$\boldsymbol{Q}$} denotes an $n\ \times\ n$ matrix 
and {$\boldsymbol{Q} = [Q_{ij}]$, $Q_{ij} = y_i y_j K(\boldsymbol{x}_i, \boldsymbol{x}_j)$}
and {$K(\boldsymbol{x}_i, \boldsymbol{x}_j)$} is a kernel value computed from a kernel function.

Kernel functions (e.g., the Gaussian kernel function~\cite{scholkopf2001learning}, the ideal kernel function~\cite{kwok2003learning}) are used to map the problem from the original data space to a higher dimensional data space. For a training set with $n$ instances, the $i^{th}$ row $\boldsymbol{\mathcal{K}}_i = \langle${$K(\boldsymbol{x}_i, \boldsymbol{x}_1)$}, {$K(\boldsymbol{x}_i, \boldsymbol{x}_2)$}, ...,
{$K(\boldsymbol{x}_i, \boldsymbol{x}_n)$}$\rangle$ of the kernel matrix corresponds to all the $n$ kernel values of the instance $\boldsymbol{x}_i$.

\subsection{The SMO algorithm} 
\label{paper:smo}
Problem~\eqref{eq:svm_dual} is a quadratic programming problem, and can be solved by many algorithms. Here, we describe a popular training algorithm, namely the Sequential Minimal Optimization (SMO) algorithm~\cite{platt1998sequential}, which is adopted in many existing SVM libraries such as LIBSVM~\cite{chang2011libsvm} and liquidSVM~\cite{steinwart2017liquidsvm}. The SMO algorithm iteratively improves the weight vector $\boldsymbol{\alpha}$ until the optimal condition of the SVM is met. The optimal condition is reflected by an \emph{optimality indicator vector} $\boldsymbol{f} = \langle f_1, f_2, ..., f_n \rangle$ where $f_i$ is the optimality indicator for the $i^{th}$ instance $\boldsymbol{x}_i$ and $f_i$ can be obtained using the following equation: $f_i = \sum_{j=1}^{n}{\alpha_j y_j K(\boldsymbol{x}_i, \boldsymbol{x}_j) - y_i}$. The SMO algorithm has the following three steps:

\textbf{Step 1}: Find two extreme training instances, denoted by $\boldsymbol{x}_{u}$ and $\boldsymbol{x}_{l}$,
which have the minimum and maximum optimality indicators, respectively.
The indexes of $\boldsymbol{x}_{u}$ and $\boldsymbol{x}_{l}$, denoted by $u$ and $l$ respectively, can be computed by the following equations \cite{fan2005working}.
\begin{equation}
\label{eq:chooseExtreme}
\begin{gathered}
    u = \argminl_{i}\{f_i| \boldsymbol{x}_i \in \mathcal{X}_{upper}\} \\
    l = \argmaxl_{i}\{\frac{(f_{u} - f_i)^2}{\eta_i} | f_{u}<f_i, \boldsymbol{x}_i \in \mathcal{X}_{lower}\}
\end{gathered}
\end{equation}
where \\
\vspace{-8pt}
$$\mathcal{X}_{upper} = \mathcal{X}_1 \cup \mathcal{X}_2 \cup \mathcal{X}_3,~\mathcal{X}_{lower} = \mathcal{X}_1 \cup \mathcal{X}_4 \cup \mathcal{X}_5$$
\vspace{-8pt}
and\\
$$\mathcal{X}_{1} = \{\boldsymbol{x}_i| \boldsymbol{x}_i \in \mathcal{X}, 0 < \alpha_i < C\}$$
$$\mathcal{X}_{2} = \{\boldsymbol{x}_i| \boldsymbol{x}_i \in \mathcal{X}, y_i = +1, \alpha_i = 0\}$$
$$\mathcal{X}_{3} = \{\boldsymbol{x}_i| \boldsymbol{x}_i \in \mathcal{X}, y_i = -1, \alpha_i = C\}$$
$$\mathcal{X}_{4} = \{\boldsymbol{x}_i| \boldsymbol{x}_i \in \mathcal{X}, y_i = +1, \alpha_i = C\}$$
$$\mathcal{X}_{5} = \{\boldsymbol{x}_i| \boldsymbol{x}_i \in \mathcal{X}, y_i = -1, \alpha_i = 0\}$$
$$\eta_i = K(\boldsymbol{x}_{u}, \boldsymbol{x}_{u}) + K(\boldsymbol{x}_{i}, \boldsymbol{x}_{i}) - 2K(\boldsymbol{x}_{u}, \boldsymbol{x}_{i})$$
\vspace{-10pt}

{$f_{u}$} and {$f_{l}$} denote the optimality indicators of {$\boldsymbol{x}_{u}$} and {$\boldsymbol{x}_{l}$}, respectively.

\textbf{Step 2}: Improve the weights of $\boldsymbol{x}_{u}$ and $\boldsymbol{x}_{l}$,
denoted by $\alpha_{u}$ and $\alpha_{l}$, by updating them as follows.
$$\alpha_{l}' = \alpha_{l} + \frac{y_{l}(f_{u} - f_{l})}{\eta},~\alpha_{u}' = \alpha_{u} + y_{l} y_{u}(\alpha_{l} - \alpha_{l}')$$
where {$\eta = K(\boldsymbol{x}_{u}, \boldsymbol{x}_{u}) + K(\boldsymbol{x}_{l}, \boldsymbol{x}_{l}) - 2K(\boldsymbol{x}_{u}, \boldsymbol{x}_{l})$}.
To guarantee the update is valid, when $\alpha_{u}'$ or $\alpha_{l}'$ exceeds the domain of $[0, C]$,
$\alpha_{u}'$ and $\alpha_{l}'$ are adjusted into the domain.

\textbf{Step 3}: Update the optimality indicators of all the training instances.
The optimality indicator $f_i$ of the instance $\boldsymbol{x}_i$ is updated to $f'_i$ using the following formula:
$$f_i' = f_i + (\alpha_{u}' - \alpha_{u})y_{u} K(\boldsymbol{x}_{u}, \boldsymbol{x}_i)\ +\ (\alpha_{l}' - \alpha_{l}) y_{l} K(\boldsymbol{x}_{l}, \boldsymbol{x}_i)$$

SMO repeats the above steps until the following condition is met.

\begin{equation}
f_{u} \ge f_{max} = max\{f_i | \boldsymbol{x}_i \in \mathcal{X}_{lower}\}
\label{eq:real_max_f}
\end{equation}

\subsection{A more recent SVM training algorithm based on SMO}
As we have mentioned in Section~\ref{paper:smo}, the SMO algorithm selects two training instances (which together form a working set) to improve the current SVM. Instead of using a working set of size two, a more recent SVM training algorithm uses a bigger working set and solves multiple subproblems of SMO in a batch~\cite{wenthundersvm18}, which is implemented in ThunderSVM\footnote{For ease of presentation, we use ThunderSVM to refer to ``the SVM training algorithm implemented in ThunderSVM''.}.

Given the training dataset, ThunderSVM does the following steps to train the SVM. First, a working set is formed with a number of instances that violate the optimality condition the most. Second, the rows of kernel values needed for the subproblems corresponding to the working set are computed. Third, the SMO algorithm is used to solve each of the subproblems. ThunderSVM repeats the above steps until the termination criterion is met.

The second step can be done by matrix multiplication, and the third step can be solved by SMO as discussed in Section~\ref{paper:smo}. Solving the first step is much more challenging. Here we elaborate the details about the first step. At each iteration when ThunderSVM updates the working set, $q$ instances in the working set will be replaced with $q$ new violating instances (e.g., $q=512$ by default in ThunderSVM). The intuition for updating the working set is to choose $q$ training instances that violate the optimality condition (cf. Inequality~\ref{eq:real_max_f}) the most, such that the current SVM can be potentially improved the most. ThunderSVM does not update the whole working set to mitigate the local optimization. ThunderSVM sorts the optimality indicators ascendingly. Then, it chooses the top $q/2$ training instances whose $y_i\alpha_i$ can be increased, and the bottom $q/2$ training instances whose $y_i\alpha_i$ can be decreased. ThunderSVM considers $y_i\alpha_i$, because of the constraints $0 \leq \alpha_i \leq C$ and $\sum_{i=1}^{n}{y_i\alpha_i} = 0$ in Problem~\eqref{eq:svm_dual}.

In summary, the training algorithm of ThunderSVM has two phases: selecting a working set from the training dataset, and solving the subproblems using SMO. Since the same row of the kernel values is often used multiple times in different iterations during the training, we can adopt a cache to store the kernel values to reuse them between the iterations.

\subsection{Existing SVM libraries and caching}
One of the key factors for the success of SVMs is that many easy-to-use libraries are available. The popular libraries include SVM$^{light}$~\cite{joachims1998making}, LIBSVM~\cite{chang2011libsvm} and ThunderSVM. SVM$^{light}$, LIBSVM and many other SVM libraries (e.g., liquidSVM) adopt the LRU strategy for kernel value caching, while ThunderSVM has not applied caching strategy. The LRU caching strategy works well for occasions with good temporal locality. However, no evidence has shown that the time of the reuse interval of kernel values is relatively small, and hence the LRU caching strategy may not be a good option for kernel value caching in the SVM training.

MASCOT~\cite{wen2014mascot} adopts a new caching strategy called LAT. When the cache is full, LAT replaces the row with the minimum row index in the kernel matrix. LAT effectively caches the last part of the kernel matrix which is stored in SSDs in MASCOT. This makes LAT prefer to cache kernel values stored in SSDs rather than those in the main memory. LAT shows better performance than LRU on MASCOT.

\section{Motivations}
\label{paper:motivations}
For ease of presentation, we call one row of kernel matrix an \emph{item}, which is the smallest unit for the replacement operation in the training. Our design of HCST is motivated by the following observations of the access pattern of the items.

\begin{figure}
\begin{center}
\includegraphics[width=2.8in]{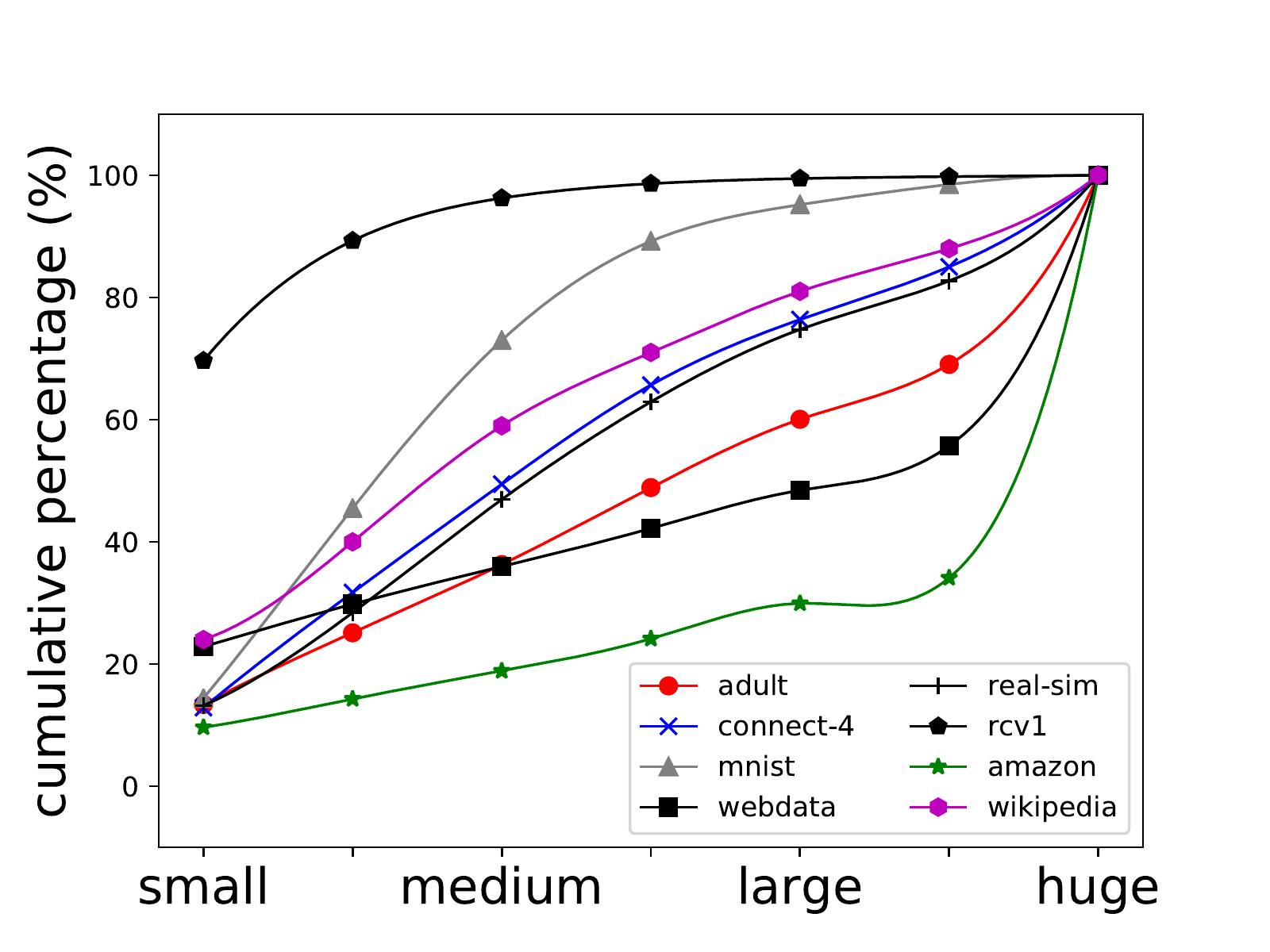}
\vspace{-15pt}
\caption{Interval of two consecutive accesses of the same item}\label{fig:interval}
\end{center}
\vspace{-20pt}
\end{figure}

\emph{Observation 1: Due to the relatively large reuse interval, LRU is not suitable for the overall training.}
Suppose the cache can store at most 5,000 items. We define \emph{reuse interval $R$} to be the number of iterations between two consecutive accesses to an item. For clarity of presentation, we divide the reuse interval $R$ to four different levels: \emph{small} ($0 < R < 5,000$), \emph{medium} ($5,000 \le R < 10,000$) , \emph{large} ($10,000 \le R < 15,000$) and \emph{huge} ($R \ge 15,000$). Figure~\ref{fig:interval} shows the cumulative percentage of different levels. LRU works well if most reuse intervals are small so that the item is still in cache when accessed again. However, as we can see from the results, except for \emph{rcv1}, the proportion of small reuse interval is very low, which means LRU is not a suitable strategy for the kernel value caching in SVM training. Wen \emph{et al}.~\cite{wen2014mascot} has proved that the items are accessed in a quasi-round-robin manner, where the selected items are unlikely to be used again in the near future of the training, which is consistent with our observation.

\begin{figure}[!t]
\captionsetup[subfloat]{farskip=2pt,captionskip=1pt}
\centering
\subfloat[adult]{\includegraphics[width=1.75in]{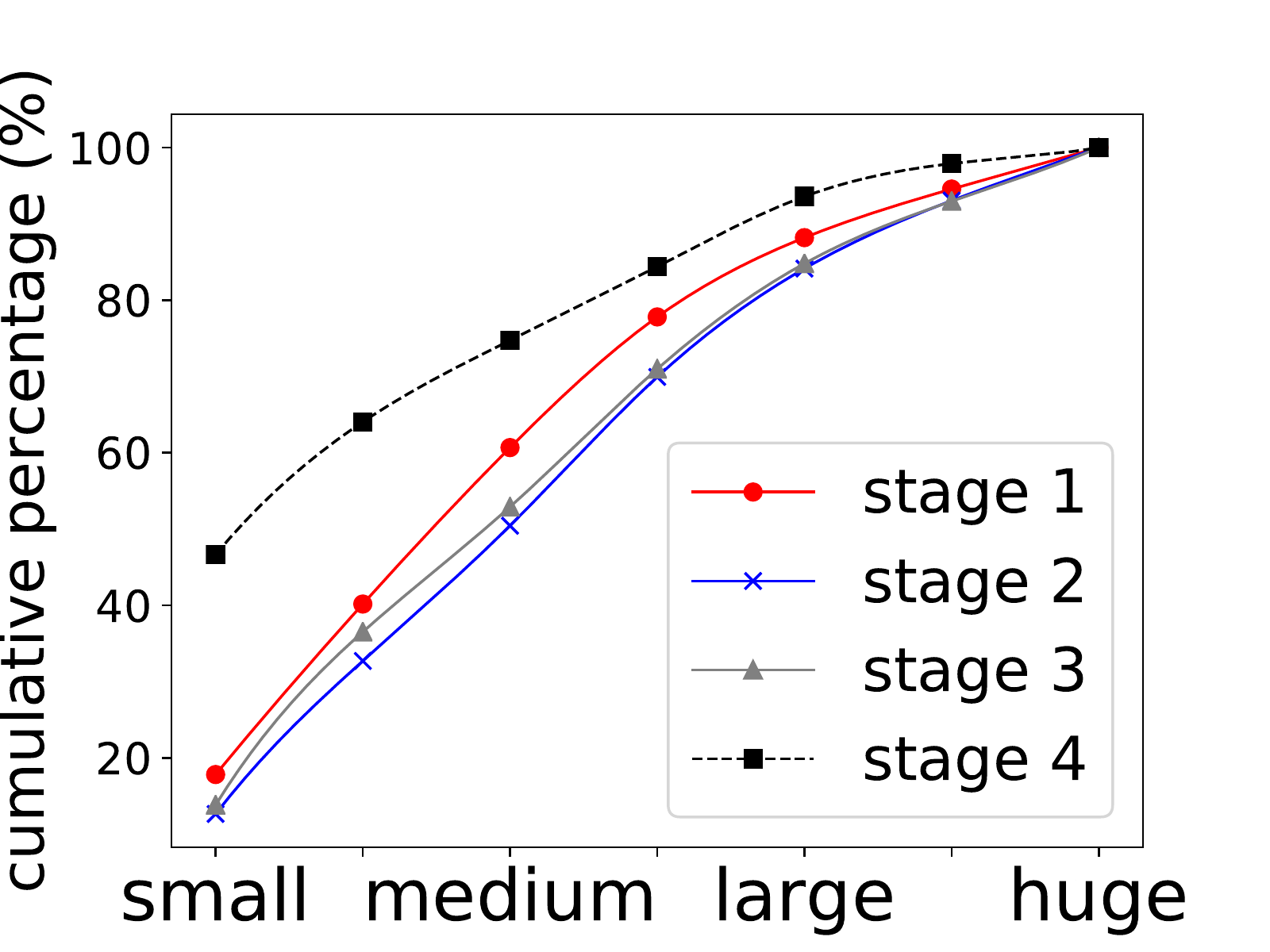}%
}
\subfloat[connect-4]{\includegraphics[width=1.75in]{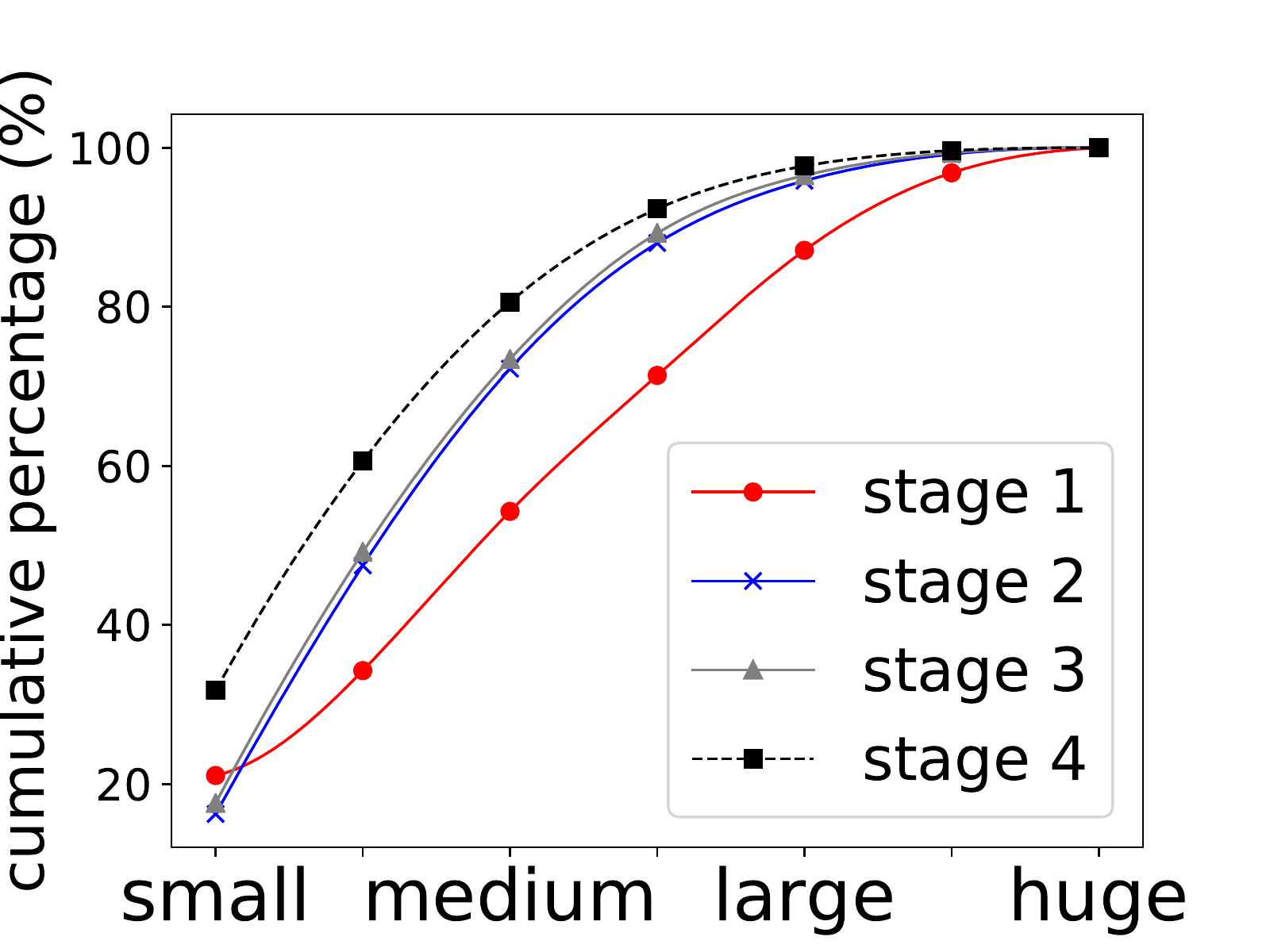}%
}

\subfloat[mnist]{\includegraphics[width=1.75in]{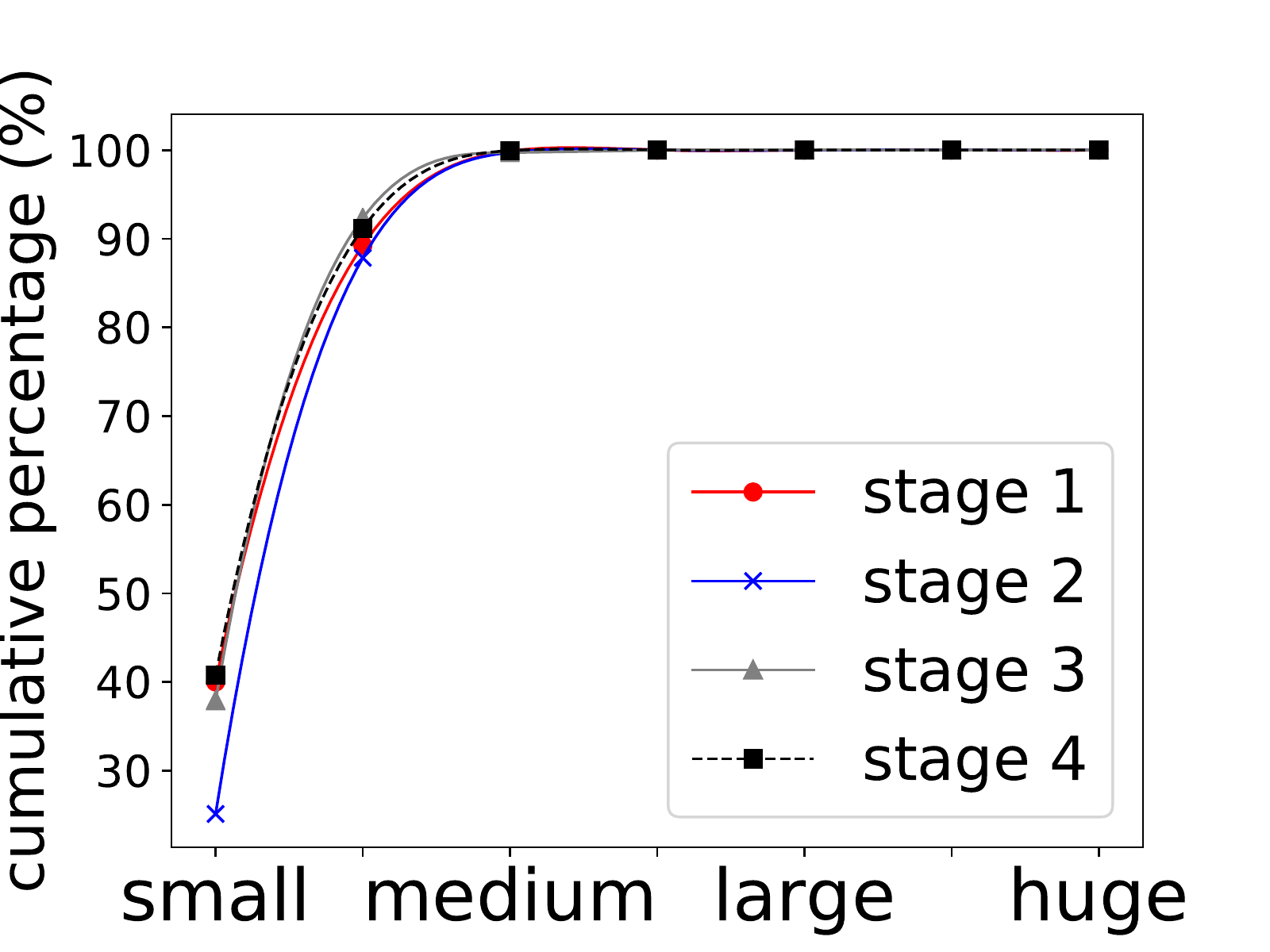}%
}
\subfloat[webdata]{\includegraphics[width=1.75in]{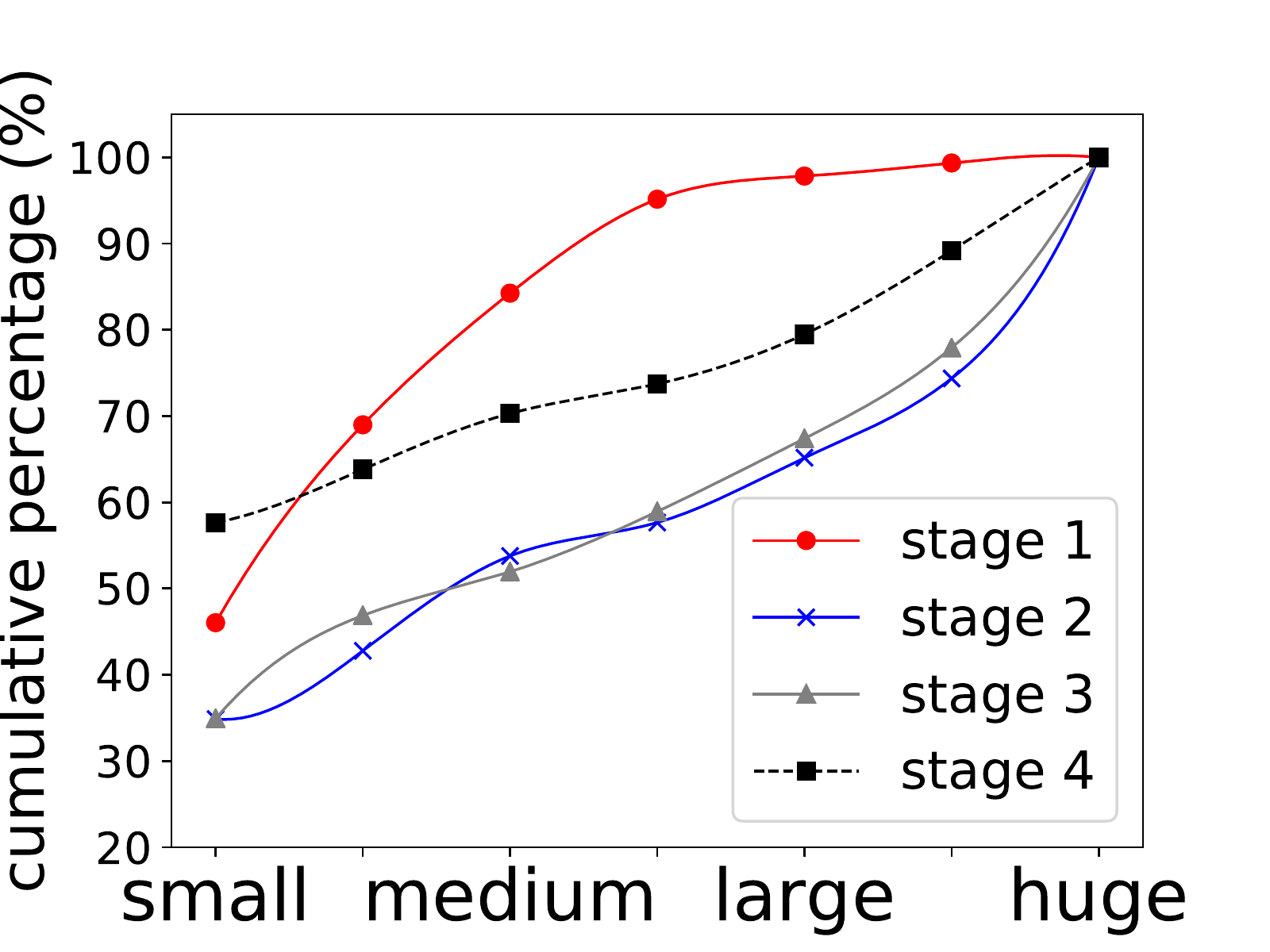}%
}

\subfloat[real-sim]{\includegraphics[width=1.75in]{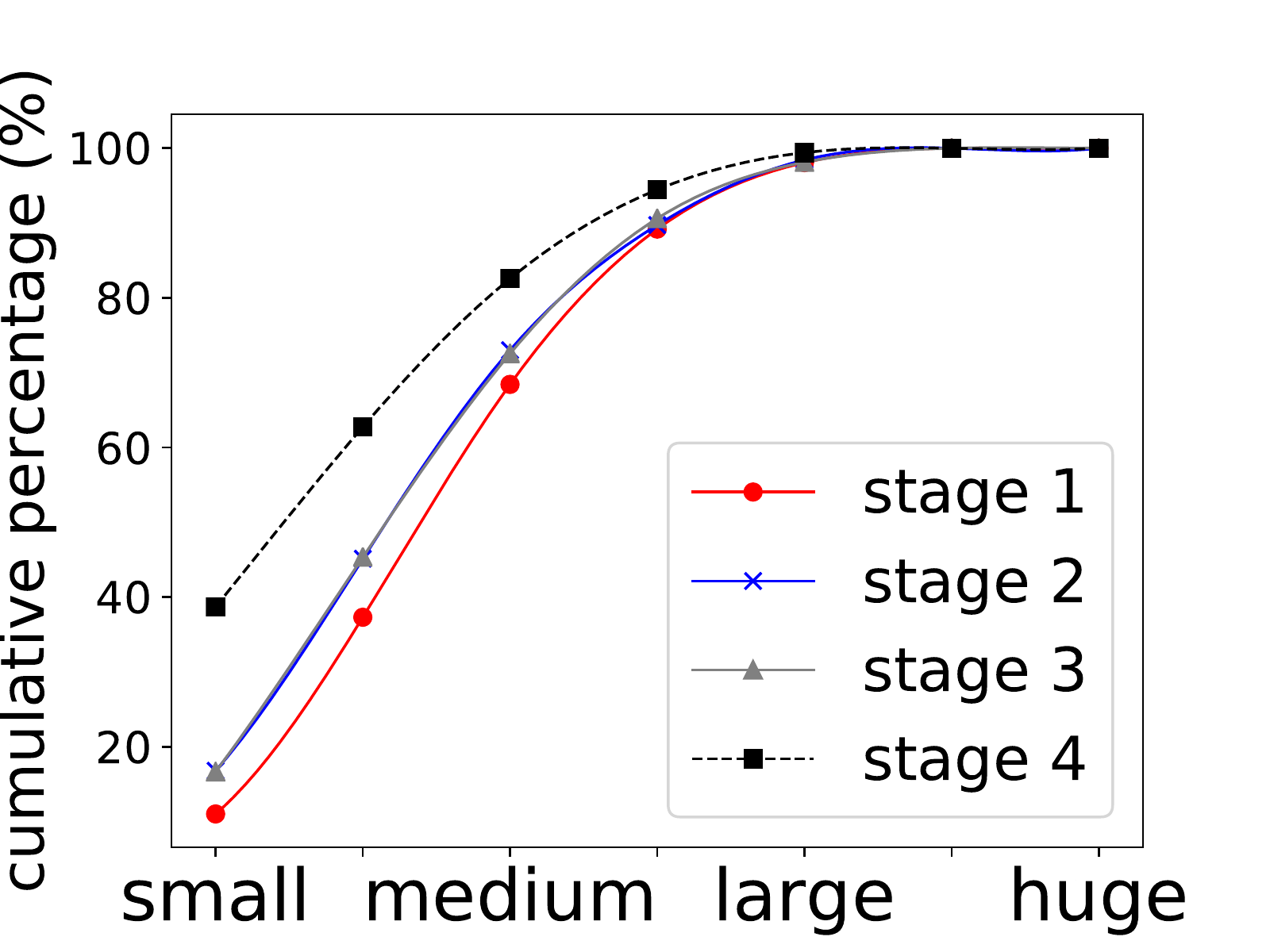}%
}
\subfloat[rcv1]{\includegraphics[width=1.75in]{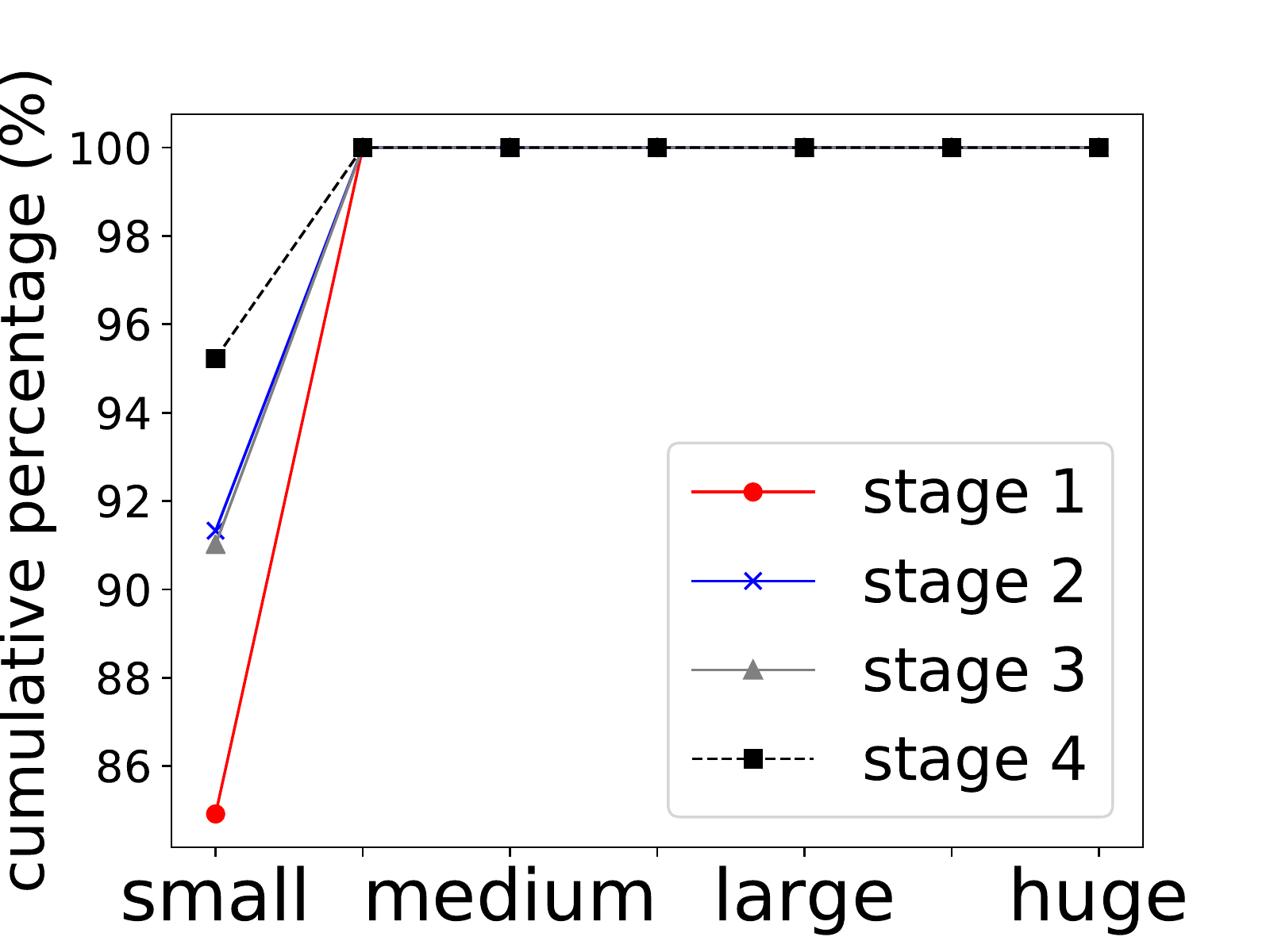}%
}

\subfloat[amazon]{\includegraphics[width=1.75in]{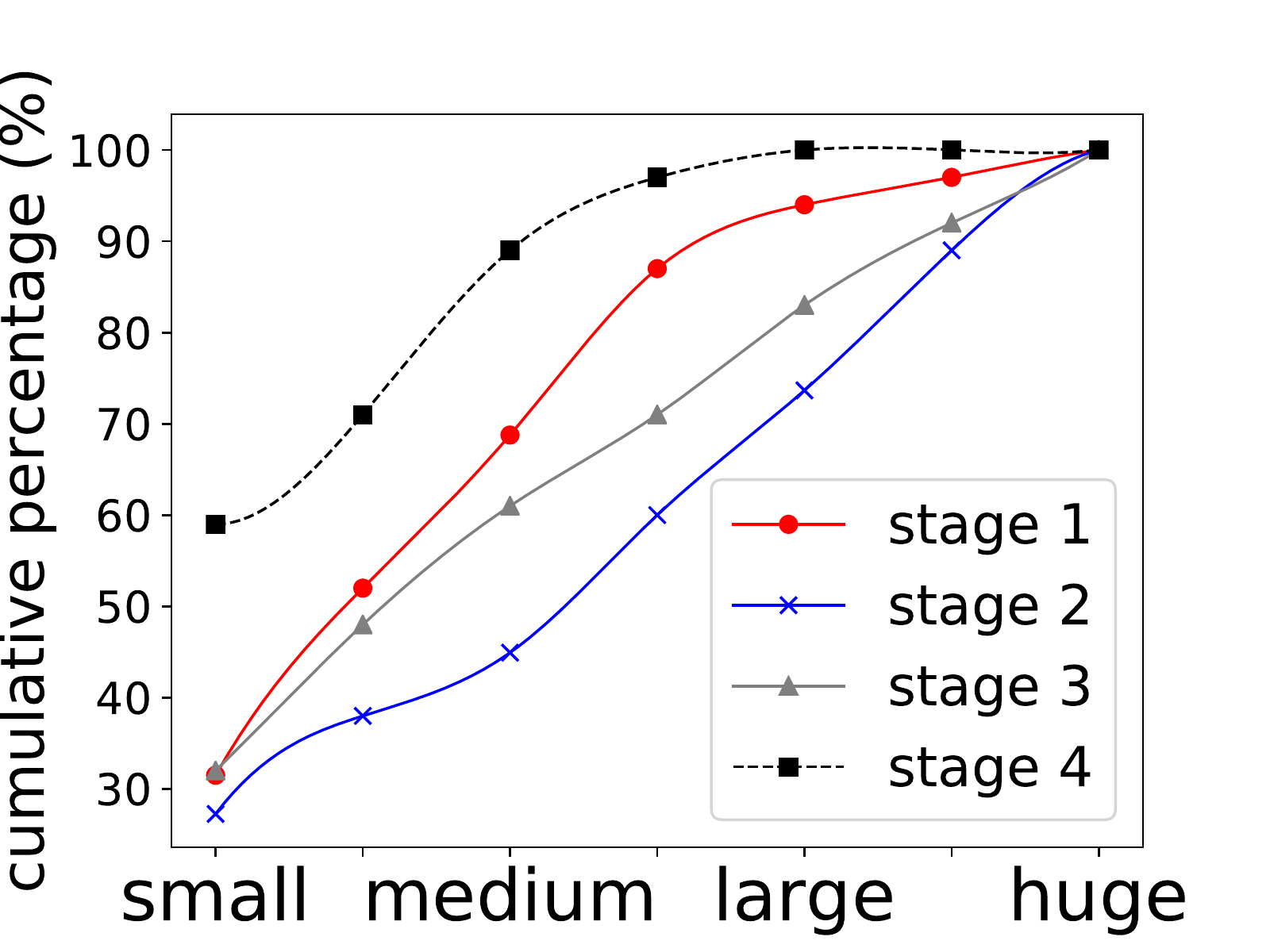}%
}
\subfloat[wikipedia]{\includegraphics[width=1.75in]{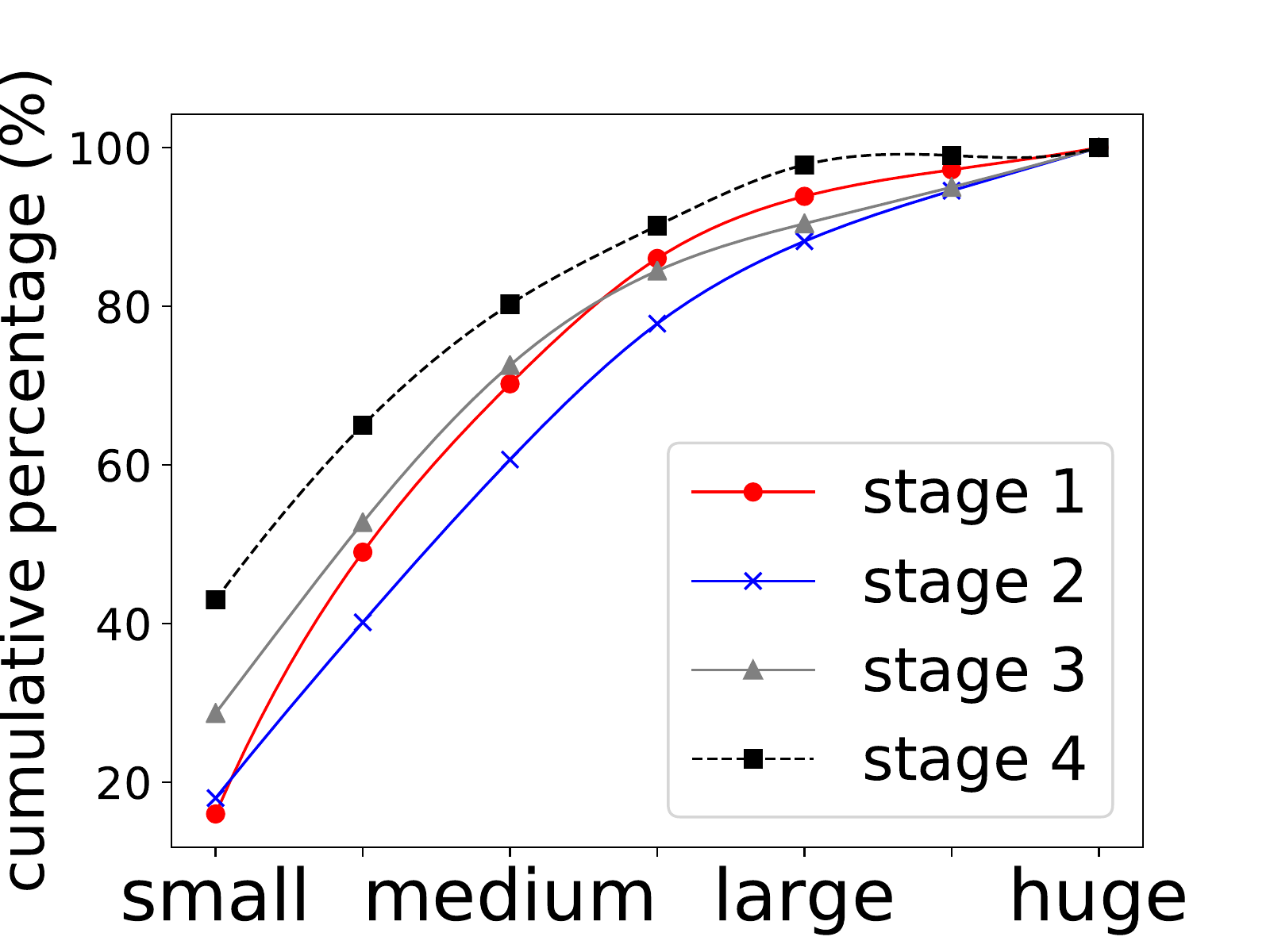}%
}
\caption{Cumulative distribution of reuse intervals in different stages}
\label{fig:reuseDisPart}
\vspace{-15pt}
\end{figure}

\emph{Observation 2: LRU may have a better performance in the late stage than the early stages of the training process, where the proportion of small reuse intervals is relatively large.}
To analyze changes in the access patterns throughout the whole SVM training, we divide the training process into four stages evenly according to the number of iterations. We have also tested different number of stages such as two and eight, and observed similar results. Figure~\ref{fig:reuseDisPart} shows the cumulative distribution of reuse intervals in different stages. We observe that there are some common features of the access patterns in the training progress. The stage 4 always has the highest proportion of small reuse intervals. Here is a scientific explanation. Compared with the early stages, the SVM training is more likely to choose the support vectors to adjust the hyperplane near the end of the training process~\cite{gu2018accelerated}. Then, the distribution of the accesses is more concentrated in the late stage. Thus, the proportion of small reuse interval is larger in the late stage, where LRU may have a good performance. So LRU can perform better in the late stage of the training.

\begin{figure}[!t]
\captionsetup[subfloat]{farskip=2pt,captionskip=1pt}
\centering
\subfloat[adult]{\includegraphics[width=1.75in]{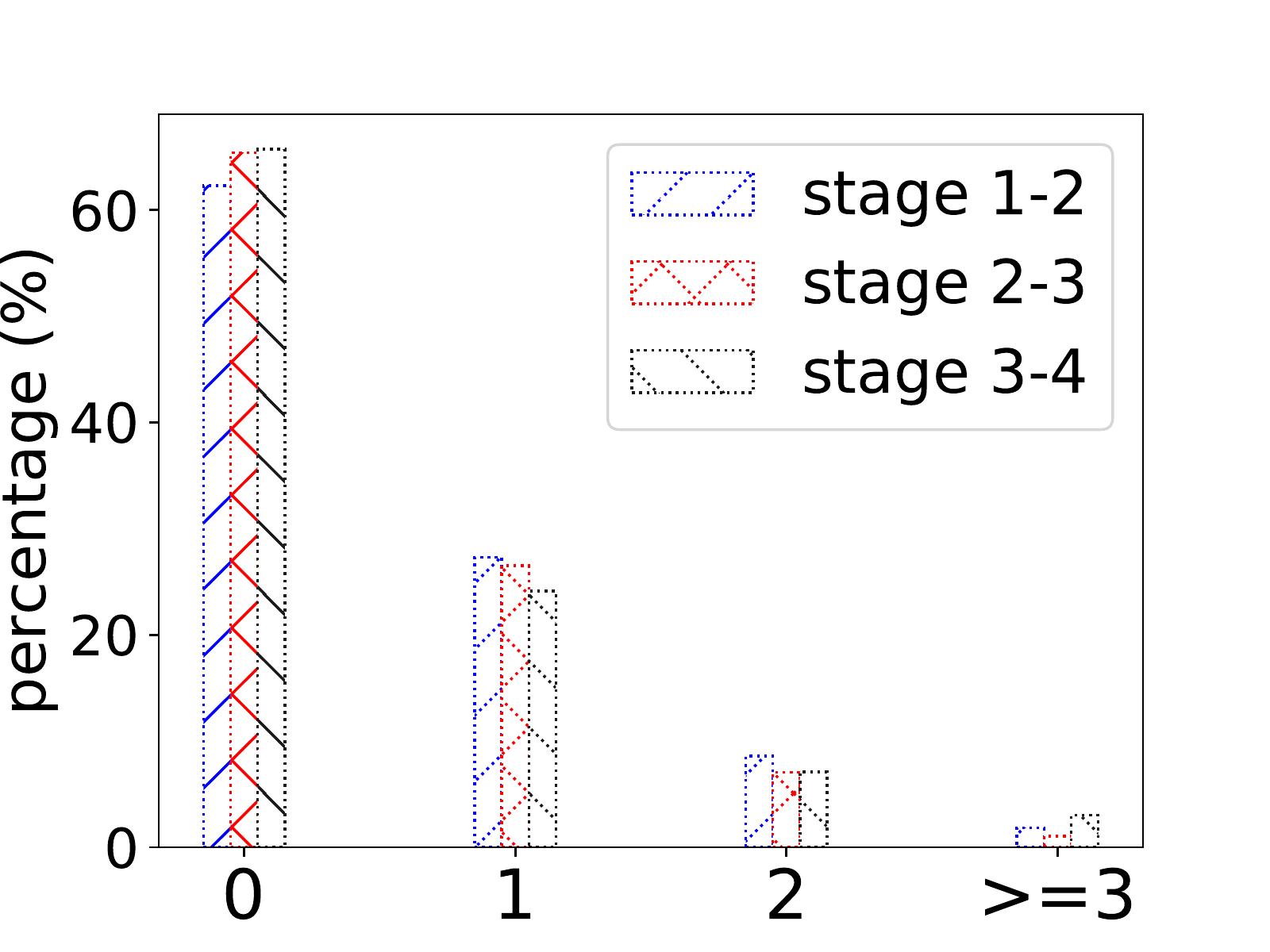}%
}
\subfloat[connect-4]{\includegraphics[width=1.75in]{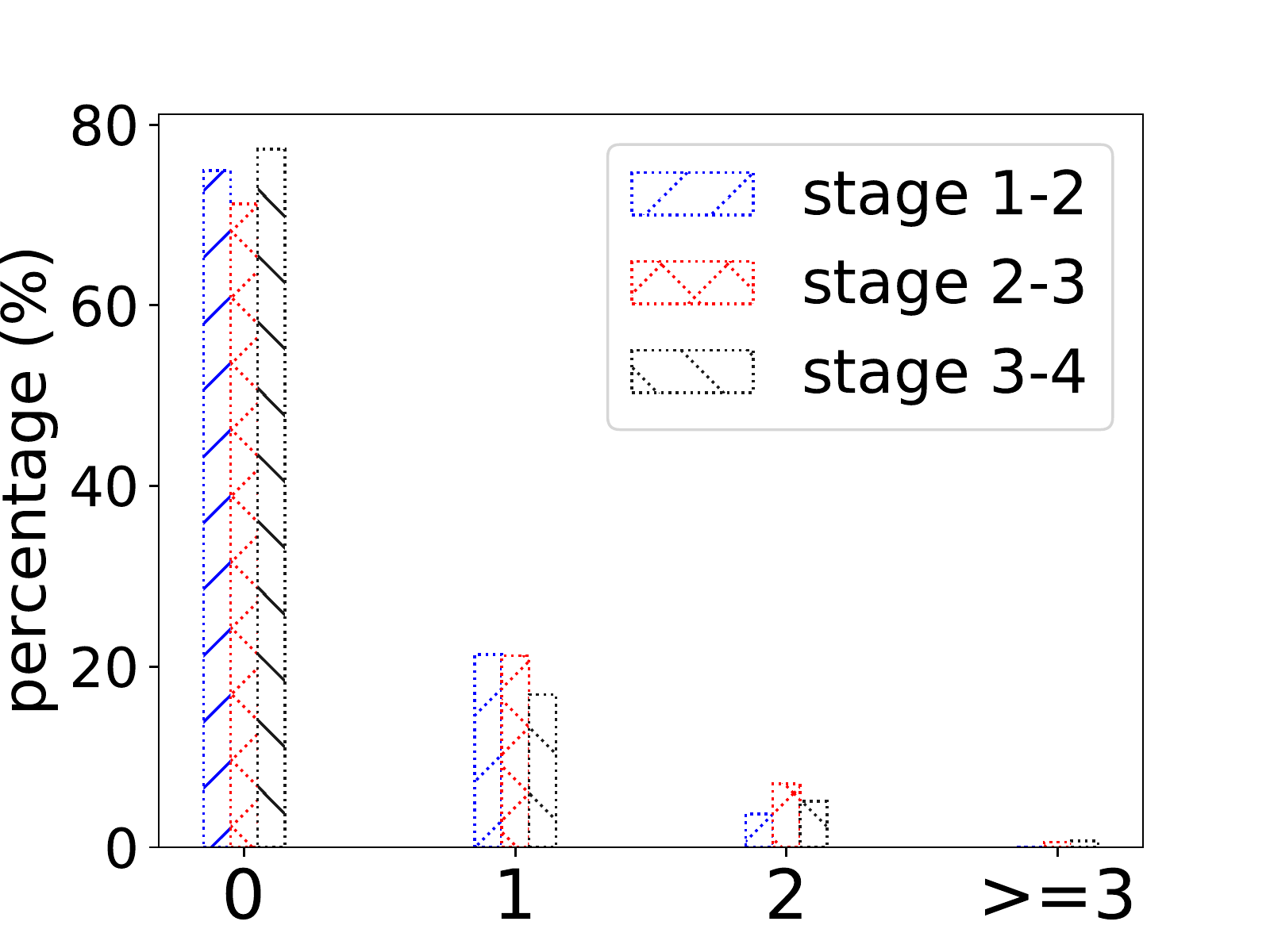}}

\subfloat[mnist]{\includegraphics[width=1.75in]{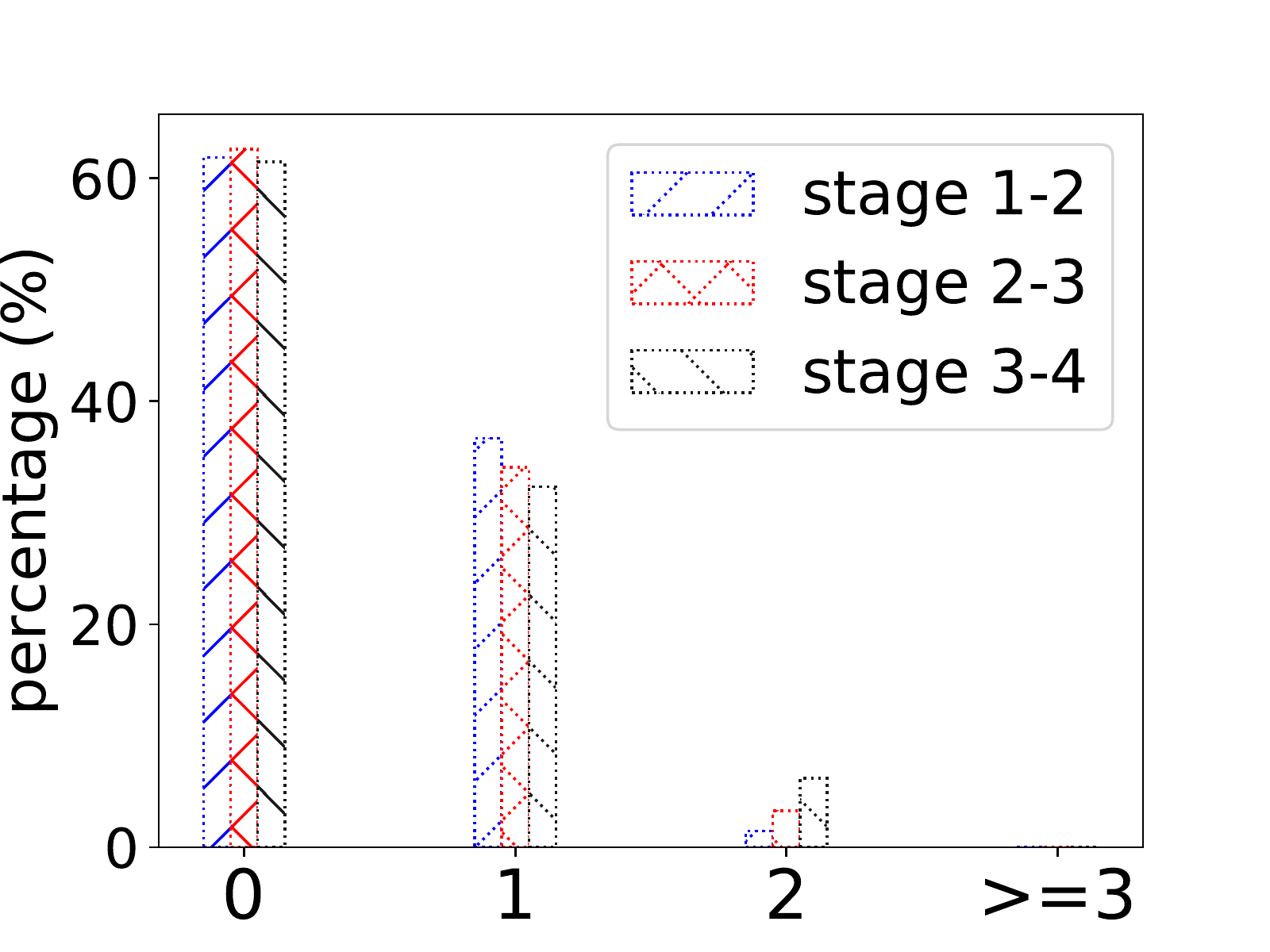}%
}
\subfloat[webdata]{\includegraphics[width=1.75in]{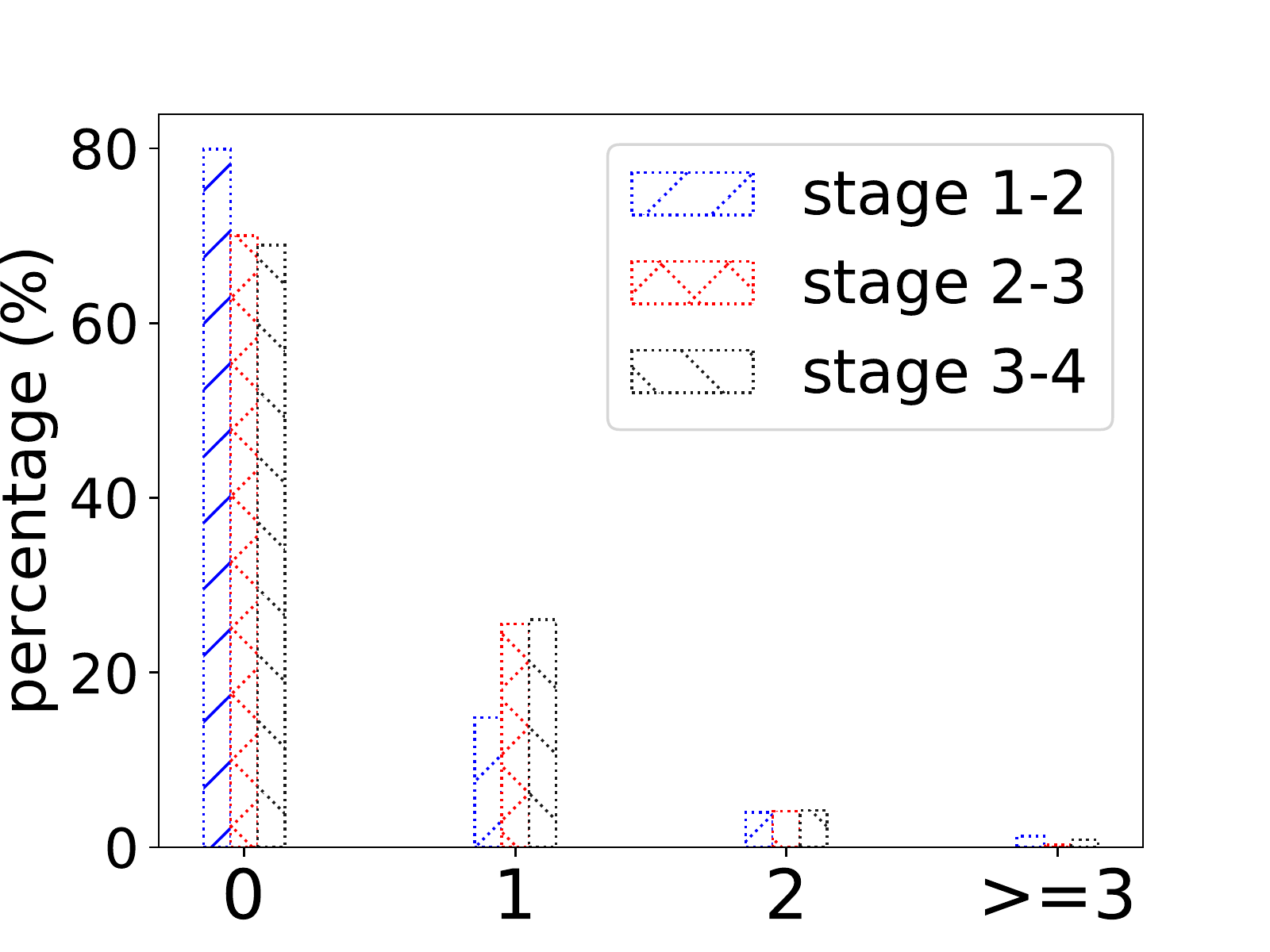}%
}

\subfloat[real-sim]{\includegraphics[width=1.75in]{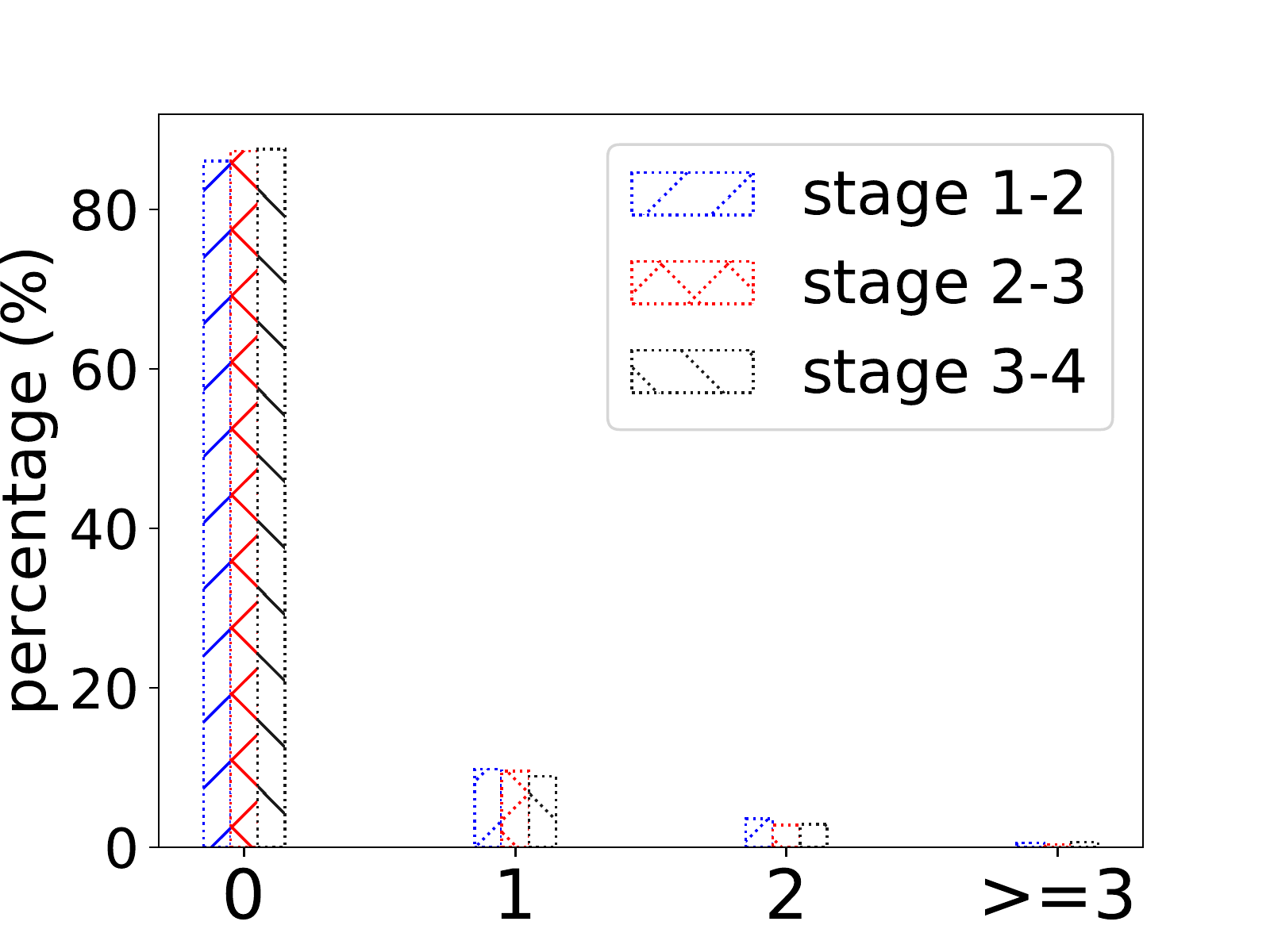}%
}
\subfloat[rcv1]{\includegraphics[width=1.75in]{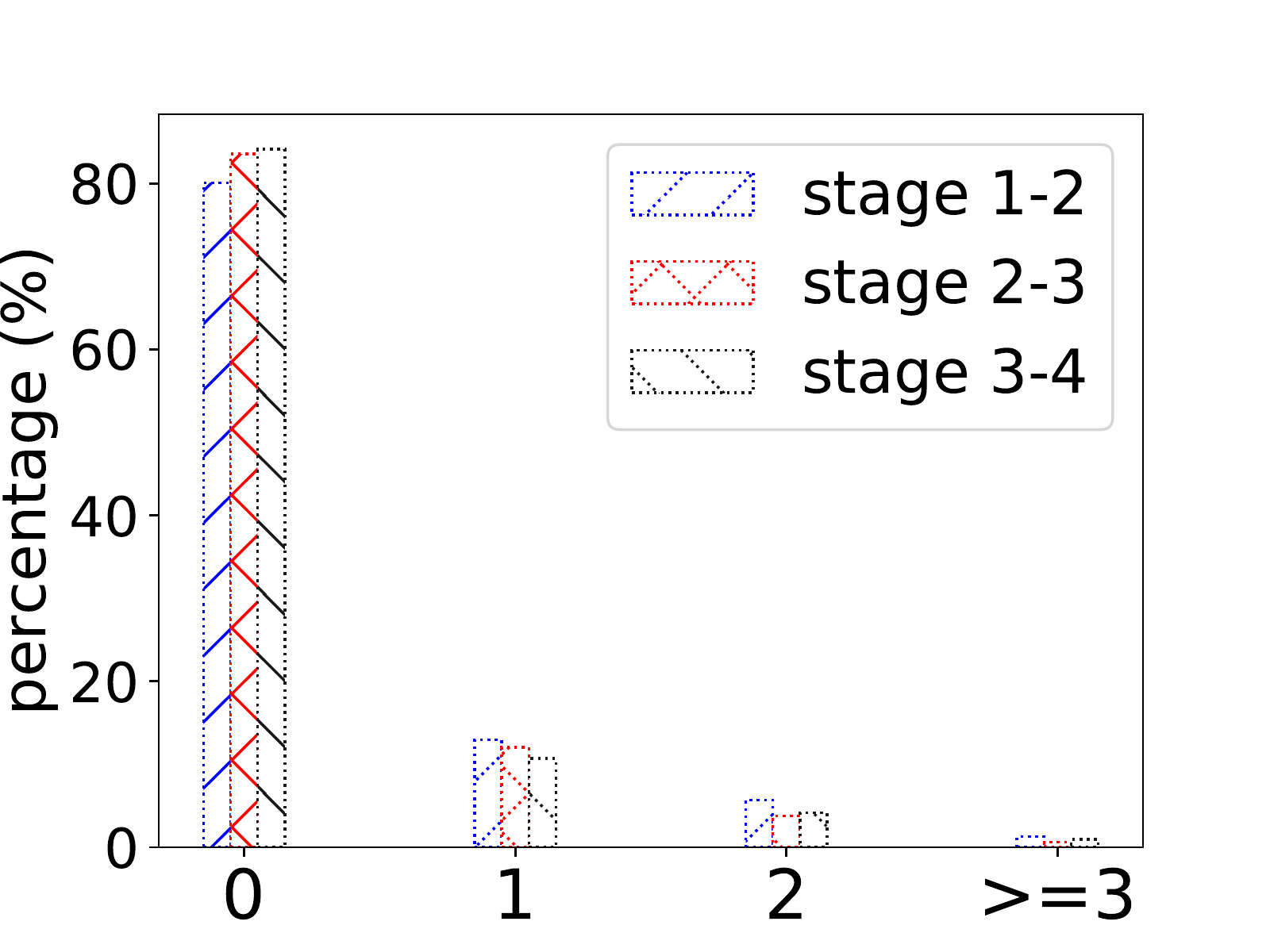}%
}

\subfloat[amazon]{\includegraphics[width=1.75in]{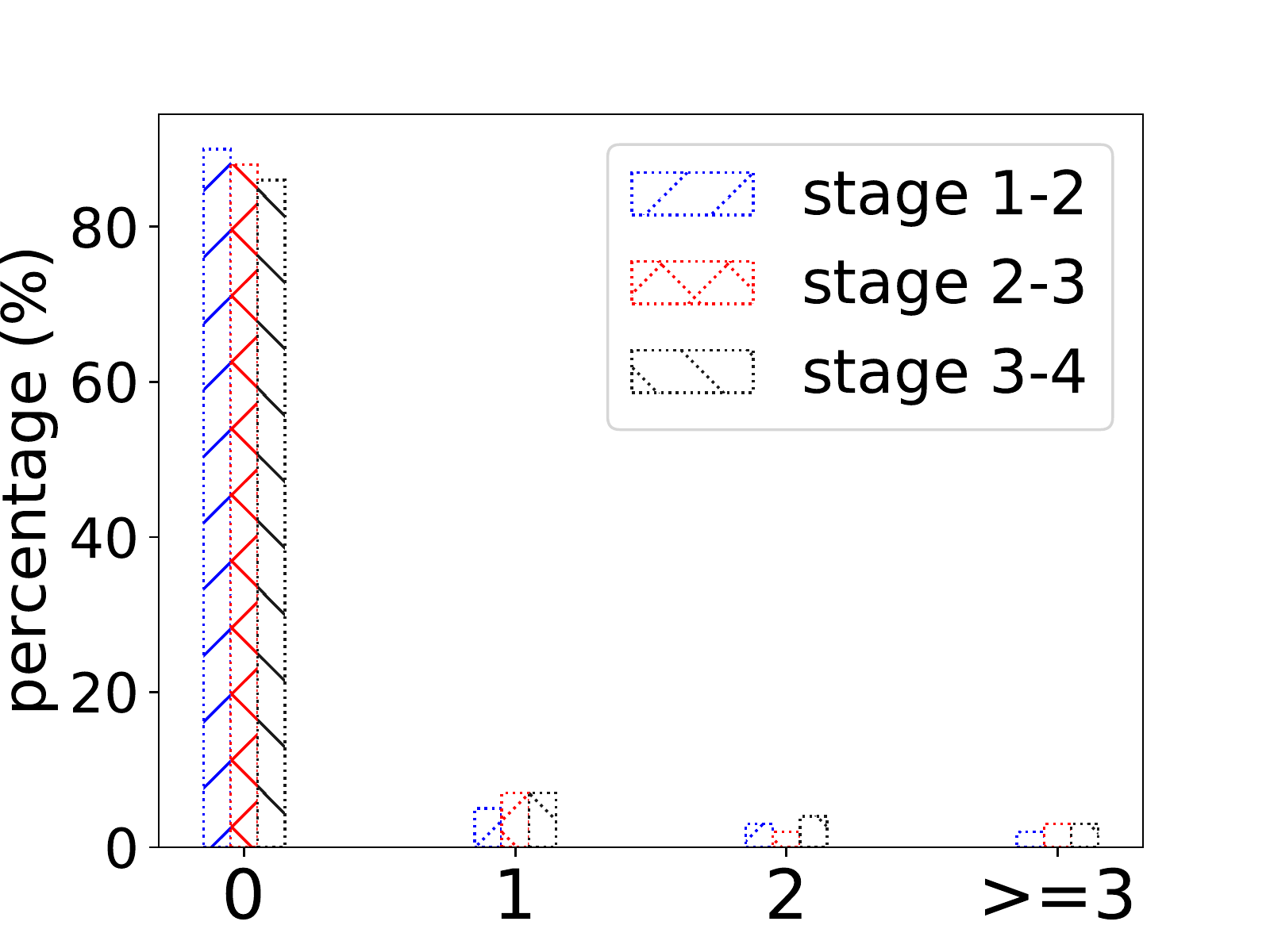}%
}
\subfloat[wikipedia]{\includegraphics[width=1.75in]{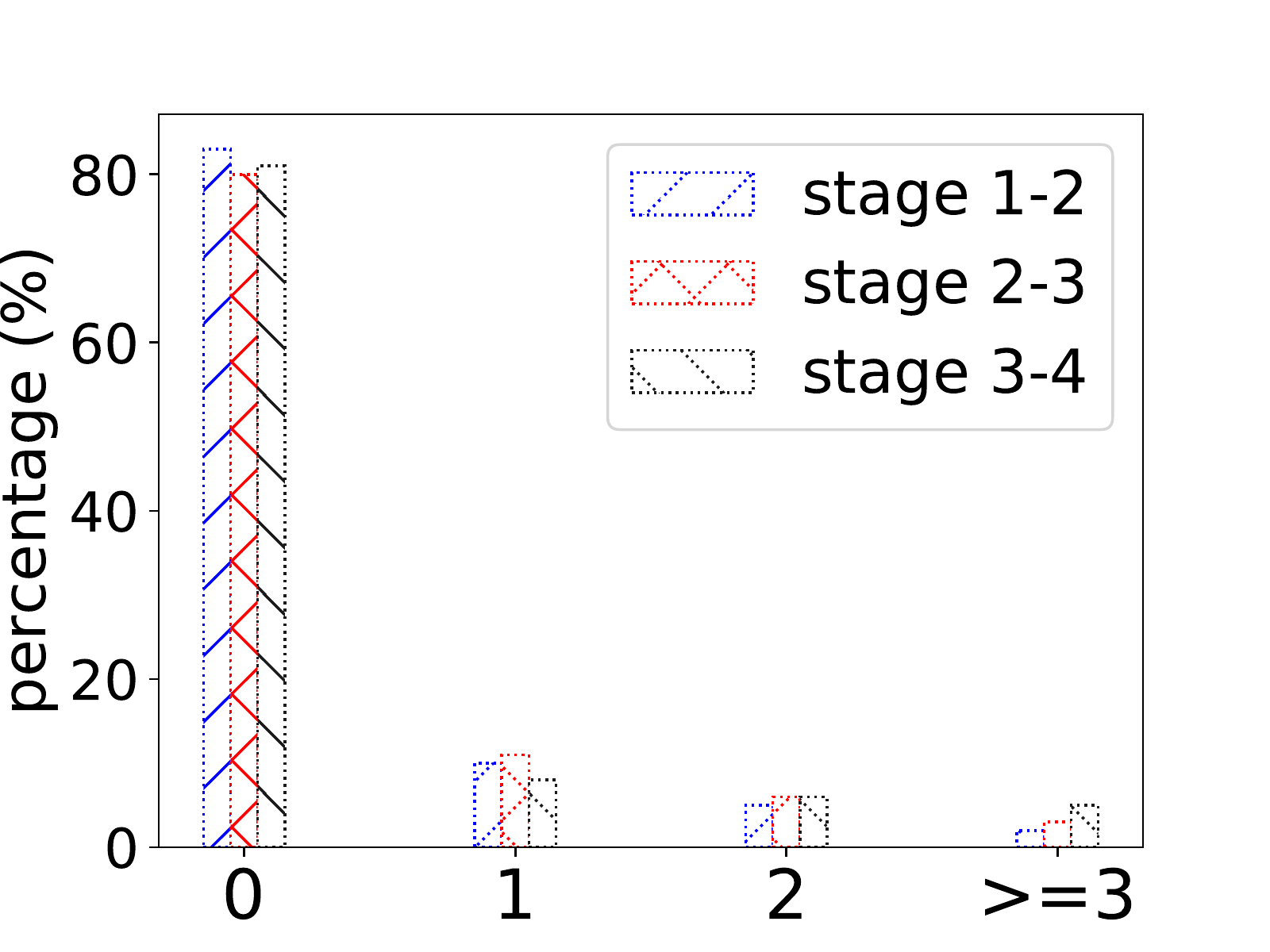}%
}
\caption{Distribution of the difference of access frequencies in different stages}
\label{fig:freDiff}
\vspace{-15pt}
\end{figure}
\emph{Observation 3: The items tend to have similar access frequency across different stages.}
In order to study the uniformity of access frequencies of the items, we calculate the difference of access frequencies of the same item between different stages. The distribution of the difference are shown in Figure~\ref{fig:freDiff}. We can find the distribution of access frequencies is quite uniform in different stages. More than 60\% of the items have no difference in the access frequencies of different stages. This observation results from the property of the SVM training. In many real world problems, most training instances are non-support vectors, which are not actively selected to update the hyperplane in the whole training process~\cite{joachims1998making}. Then, for most items, the access frequencies across different stages tend to be low and close. Based on this characteristic, items with higher access frequency should be cached since they are likely to be accessed with a higher frequency in the subsequent training progress. A classic algorithm for frequency based cache replacement is LFU. However, the LFU strategy always caches the newly generated item and replaces the least frequently used item in cache, even though the item in cache has a higher frequency than the new item, which is not good enough for caching items with higher access frequencies. That is why we propose to enhance LFU with EFU.

\section{The HCST strategy}
\label{paper:hcst}

\subsection{An Overview of HCST}
\begin{figure}
\begin{center}
\includegraphics[width=3.2in]{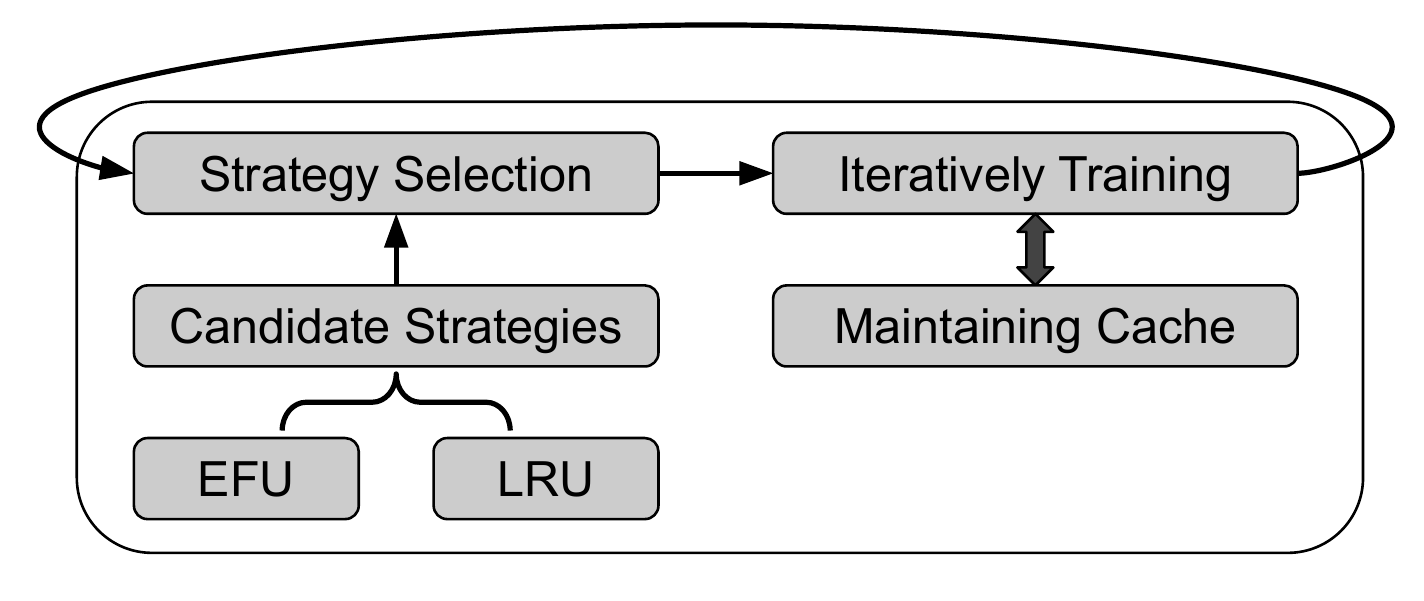}
\vspace{-5pt}
\caption{The process of the SVM training with HCST}\label{fig:overStru}
\end{center}
\vspace{-15pt}
\end{figure}

The structure of the SVM training with HCST is shown in Figure~\ref{fig:overStru}. There are mainly two components of HCST: the candidate strategies and the strategy selection. In the iterative process of the SVM training, HCST selects the better caching strategy from candidate strategies and applies it until the next selection. The main aim of HCST is to improve the hit ratio of items in the entire training process.

\subsubsection{Candidate strategies}
According to the observations in Section~\ref{paper:motivations}, we use EFU and LRU as our candidate strategies. EFU is our proposed caching strategy, which aims to cache the items with higher access frequencies. We will introduce the details of EFU in Section~\ref{sec:efu}. LRU is a classic caching strategy, which aims to cache the items with recently time used. By splitting the training process to different stages, as we claimed in Section~\ref{paper:motivations}, we have two key findings. One is that the items with higher access frequency are more likely to also be accessed more times in the next stage, which is fully utilized by EFU. The other one is that the reuse interval of items is likely to be smaller in the late stage of the training process, which makes LRU a potentially suitable strategy. By making full use of these two features, we propose the EFU strategy and use it and LRU as our candidate strategies.

\subsubsection{Strategy selection}
At first we adopt the EFU strategy, which appears to have a good performance in the overall training. During the training, after every fixed number of iterations, we make a selection from the candidate strategies based on collected statistics. Here we use the number of cache hits as the criterion. If the number of cache hits of EFU is bigger than that of LRU in the current stage, we adopt EFU in the next stage, otherwise we adopt LRU. In this way we can switch to LRU timely if LRU already works better than EFU. In addition, due to the jitter of the number of hits using LRU, there are some cases we switch to LRU prematurely. This deficiency is generally fine, since HCST can switch back to EFU quickly in the next comparisons. In general, HCST is pure EFU if no switch happens or piecewise EFU and LRU if any switch happens in the training process.

\subsection{The EFU strategy}
\label{sec:efu}
\begin{figure}
\begin{center}
\includegraphics[width=3.2in]{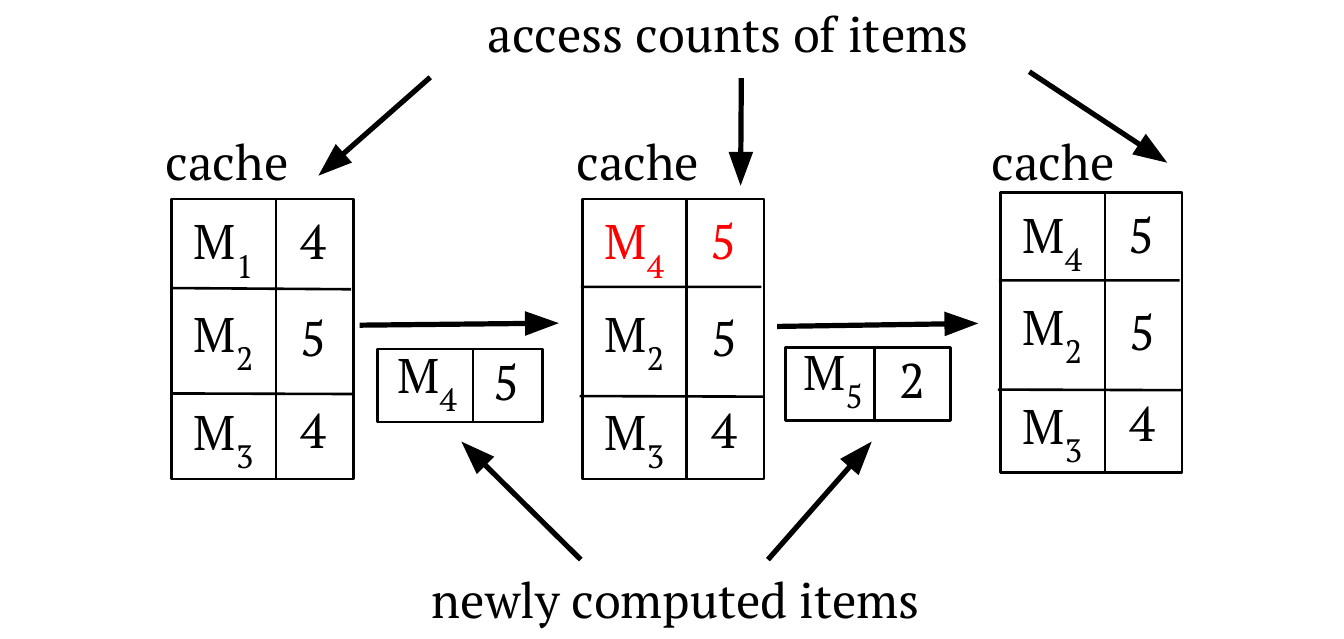}
\vspace{-5pt}
\caption{A running example of EFU}\label{fig:efuexample}
\end{center}
\vspace{-15pt}
\end{figure}

The key ideas of EFU are described as follows. We maintain a counter for each item of the kernel matrix to record the access frequency. When a new item is computed and the cache is already full, we first decide whether it will be added into the cache. If all the items in the cache have higher accumulated access frequencies than the new item, EFU will not cache it. Otherwise, EFU replaces the item in the cache with a lower access frequency than this newly computed item. In this way, we can always store the items which have higher frequency of usage currently in the SVM training. Figure~\ref{fig:efuexample} shows a running example of EFU. Suppose the cache stores 3 items: $M_1$, $M_2$ and $M_3$. These items have been accessed 4 times, 5 times and 4 times, respectively. When a newly computed item $M_4$ comes, EFU finds the item $M_1$ which has a lower access frequency than $M_4$ and replaces it. When another newly computed item $M_5$ comes, EFU does not do replacement since no item in the cache has a lower access frequency. 

The LFU strategy always replaces the least frequently used item, even though it has a higher access frequency than the new item, while the EFU strategy can avoid the problem. EFU is more suitable to store the items with higher access frequencies compared with LFU.

\subsection{The HCST strategy}
\label{paper:HCST}
As we have shown in Figure~\ref{fig:reuseDisPart}, the proportion of small reuse distance differs clearly in the late stage of the training process. There may be some cases where the performance of LRU is better than EFU in the late stages as the training processes. To handle it, we implement a dynamic strategy called HCST which allows the caching strategy to switch between EFU and LRU. At the beginning of training, we adopt EFU. In regular stages during the training, we compare the number of cache hits if adopted LRU ($H_{LRU}$) with the number of cache hits if adopted EFU ($H_{EFU}$). If we found $H_{LRU}$ is bigger than $H_{EFU}$, we use LRU for the next stage, otherwise using EFU. As we have discussed in Section~\ref{paper:motivations}, the performance of LRU strategy can be better in the late stage of the SVM training, so it is reasonable to switch the strategy to LRU if the hit number of LRU is already bigger than EFU. Furthermore, to reduce the bad effect of the premature switching from EFU to LRU, which may be caused by the instability of the number of cache hits using LRU, HCST can switch back to EFU timely in the next comparison.

In practice, it is challenging to know the exact number of cache hits of both strategies since we only can adopt one strategy at one time. However, we can estimate the number of cache hits based on the features of the cache strategy. For ease of presentation, we call the time we compare $H_{LRU}$ with $H_{EFU}$ as a checkpoint. Suppose the cache size is $s$ which means at most $s$ items can be cached.

Assume we are using EFU before a checkpoint $T_1$. We use a counter to record the number of cache hits $H_{hit}$ after $T_1$. So $H_{hit}$ is the exact $H_{EFU}$ when we arrived the next checkpoint $T_2$. We use another counter to record the number of accesses $H_{s}$ whose reuse interval is smaller than $s$ after $T_1$. Note that these accesses can all yield cache hits if using LRU. So we use $H_{s}$ as the approximate $H_{LRU}$ when we arrived $T_2$. If $H_{s}$ is not bigger than $H_{hit}$ in $T_2$, we do not change the strategy and do the same operations for the next stage.

If $H_{s}$ is bigger than $H_{hit}$, we change the strategy to LRU and save the value $H_{hit}$. When using LRU, we still use the counter to record the number of cache hits $H_{hit'}$ after checkpoint $T_2$, which is the exact $H_{LRU}$. Since the distribution of access frequencies is quite evenly as we describe in Section~\ref{paper:motivations}, the hit ratio of EFU should be close between different stages. So we use $H_{hit}$ as the approximate $H_{EFU}$. In the next checkpoint $T_3$, we compare $H_{hit'}$ with $H_{hit}$. If $H_{hit'}$ is smaller than $H_{hit}$, we switch the strategy back to EFU. Otherwise we do not change the strategy. Algorithm~\ref{alg:HCST} shows the training process with the HCST strategy.

\begin{algorithm}
\KwIn{The training dataset}
\KwOut{The SVM model}
Adopt EFU at the beginning;

\While {the optimality condition of SMO is not met}{
    \If {is a checkpoint}{
    \If {is using EFU}{
    Get the number of cache hits with EFU ($H_{hit}$);
    
    Estimate the number of cache hits with LRU ($H_{s}$);

    \If {$H_{hit} < H_{s}$}{
    Save the value $H_{hit}$;
    
    Switch to LRU;
    }
    
    }
    \Else{
    Get the number of cache hits with LRU ($H_{hit'}$);
    
    Estimate the number of cache hits with EFU ($H_{hit}$);
    
    \If{$H_{hit'} < H_{hit}$}{
    Switch to EFU;
    }
    }
    }
    Training;
}
\caption{Pseudo of the training process with HCST}
\label{alg:HCST}
\end{algorithm}

Overall, we can get the exact number of cache hits with the strategy being used and estimate the approximate number of cache hits if adopted the other strategy. By comparing these two values, the dynamic selection of the strategies can make efficient use of the characteristics we observed for the access patterns of the items.
 
\subsection{Optimization for multi-output learning tasks}
By adopting the ``one-vs-all'' scheme, we can transform a multi-output learning task into a set of independent single-output learning tasks~\cite{li2013active}. Then. we train a solver for each single-output learning task. Since each solver can utilize the CPU resources well, the solvers are trained in a sequential manner. Although the training of each solver is independent, they use the same training instances. To exploit this property, we can reuse the kernel values between different solvers. Instead of adopting individual cache for each solver, we use one shared cache during the whole training process for the multi-output tasks. Thus, after a solver finishes the training, the kernel values in the cache can still be reused in the next solver.

In summary, there are two levels of reuses of the kernel values in the multi-output learning tasks. One is the reuse of the kernel values between different iterations of a solver (iteration-level reuse). The other one is the reuse of the kernel values between different solvers of the training process (solver-level reuse). Compared with the original HCST, which only exploits the iteration-level reuse, the solver-level reuse can fully utilize the characteristic of the multi-output learning task and has no extra overhead. As we will show in Section~\ref{paper:solver-reuse}, our technique can reduce the training time significantly.
 
\subsection{Parallel implementation of the replacement operations}
\label{paper:paraReplace}
\begin{figure}
\begin{center}
\includegraphics[width=2.8in]{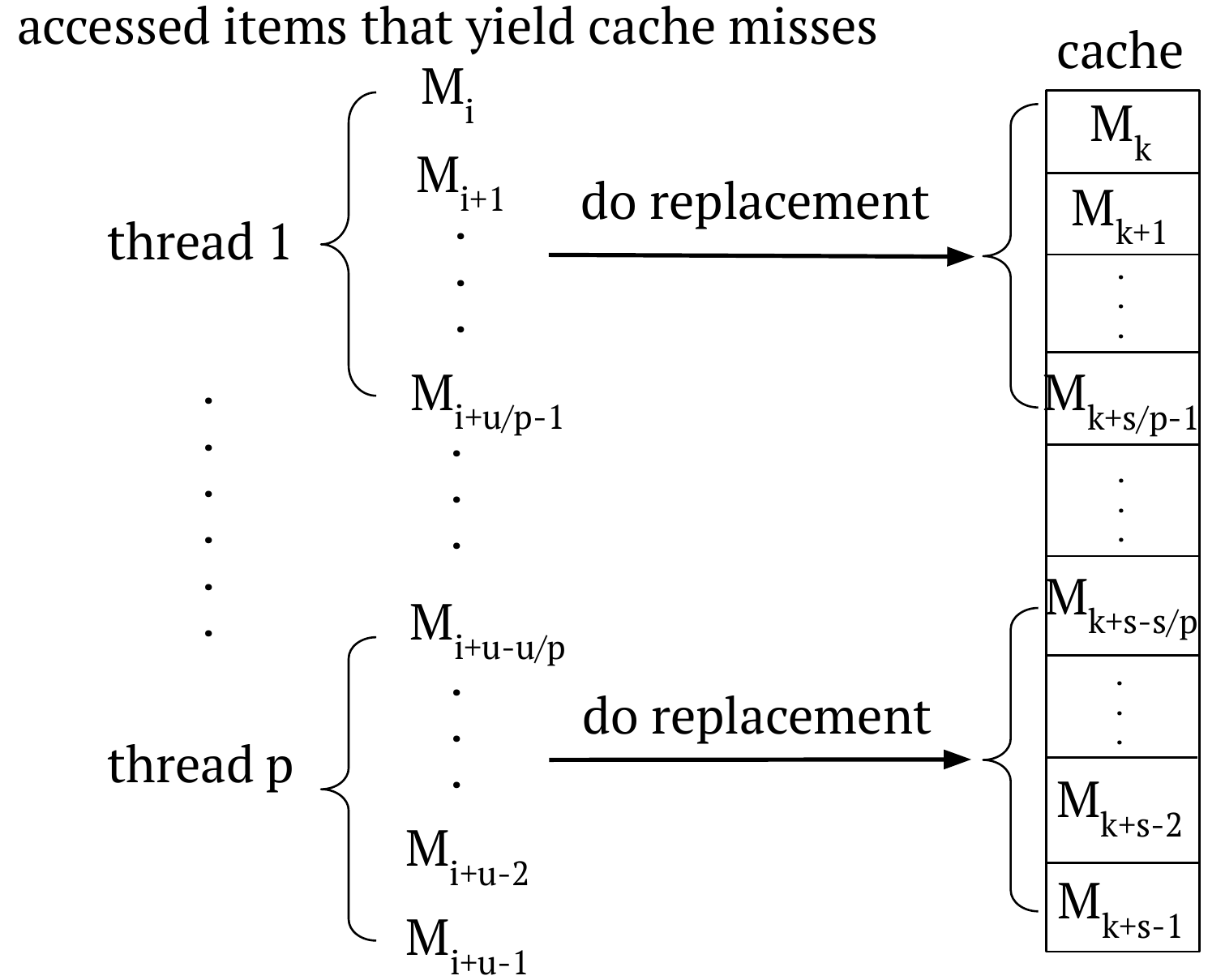}
\caption{The parallel replacement operations}\label{fig:pararep}
\end{center}
\vspace{-20pt}
\end{figure}

Since ThunderSVM accesses $q$ items on each iteration, it is time-consuming to do replacement one by one. To reduce the cost of caching, we implement the parallel replacement operation, which is shown in Figure~\ref{fig:pararep}. Suppose we have $p$ threads and there are $u$ items that yield cache misses among accessed $q$ items. We evenly divide these $u$ items into $p$ groups, and each group is assigned to one thread. To avoid conflicts in the cache when multiple threads perform the replacement, we also divide the cache into $p$ parts, and each part is assigned to one thread. Each thread traverses its corresponding part to find an eligible item to replace. Since each thread performs replacement operations for the same number of items (i.e., $u/p$) and has the same size of cache space (i.e., $s/p$) to traverse, the workload of each thread is balanced. By using multiple threads, we can reduce the cost of the replacement significantly, as we will show in Section~\ref{sec:factor}.

\subsection{Efficiency analysis}
Here we analyze the theoretical efficiency of the HCST strategy. We suppose the cardinality of the dataset is $n$ and the dimension of the dataset is $d$. ThunderSVM accesses $q$ items on each iteration.

\subsubsection{Time complexity of the replacement}
Suppose the cache size is $s$. We need at most $\mathcal{O}(s)$ to traverse on the cache to find an item that meets the requirements of the cache strategy to replace. On each iteration, there are at most $q$ items to be replaced. So the time complexity of our caching strategy in one iteration is $\mathcal{O}(qs)$.

\textit{Time complexity improvement by parallelism}: Suppose the number of threads is $p$. If we perform the replacement in parallel, we only need $\mathcal{O}(s/p)$ to find an item to replace. Additionally, each thread only needs to handle at most $q/p$ items. The time complexity of HCST in one iteration is $\mathcal{O}(qs/p^2)$, which is reduced by $p^2$ times compared with performing replacement in serial. This property makes HCST appealing in the context of parallel computing.

\textit{Overhead of the dynamic selection}: Although the HCST strategy needs a selection operation in the training process, the cost of it is very low. To get the hit number of EFU and LRU, we need maintain counters for each item as we have discussed in Section~\ref{paper:HCST}. We need $\mathcal{O}(q)$ to update these counters in each iteration, where $q$ is much smaller than $n$. The comparison between the numbers of cache hits can be done in $\mathcal{O}(1)$. The overhead of the dynamic selection is $\mathcal{O}(q)$ in each iteration, which can be ignored compared with the time complexity of the training.

\subsubsection{Memory consumption of the HCST strategy}
To enable the switch between EFU and LRU, we need maintain two counters for each item to record the access frequency and the recently used time. Suppose each integer and float use 4 bytes to store. So we need $8n$ bytes for all counters. The extra cost of memory is very small compared with the size of the dataset, which costs $4nd$ bytes to store.

\subsubsection{Benefit of caching}
We can compute the benefit of caching theoretically for the SVM training. Suppose the number of floating point operations per second (i.e., flops) of CPU is $l$. Suppose we use the Gaussian kernel for the SVM training. The Gaussian kernel is defined as follows. 
$$    K(\boldsymbol{x}_i, \boldsymbol{x}_j)  = exp\{-\gamma(||\boldsymbol{x}_i||^2+||\boldsymbol{x}_j||^2 - 2\boldsymbol{x}_i^T\boldsymbol{x}_j)\}$$

To compute an item (i.e., a row of the kernel matrix), we need to compute the inner products of all the $n$ training instances (i.e., $n$ vectors of $1 \times d$ dimensions), a multiplication of a $1 \times d$ vector and a $d \times n$ matrix (i.e., $\boldsymbol{x}_i$ and the whole training dataset). Hunger~\cite{hunger2005floating} showed the flops of the dot product computation is $(2d - 1)n$. Since inner products are computed only once at the beginning of the SVM training, we do not count them in the cost. Ignoring the floating operations to compute the exponential formula, we can get the cost for computing an item as $T_o = (2d-1)n/l$. Suppose the number of cache hits is $h$. The time of kernel value computation saved by the caching strategy $    T_s = h(2d-1)n/l$.

Let the number of replacement operations be $u$. To simplify the model, we assume the sequential copy bandwidth of memory is a constant $b$. Each kernel value is 4 bytes. The size of an item is $4n$ bytes. So we can compute the time of replacement in cache as $T_c = 4un/b$.

Ignoring other costs, we can get the benefit of caching strategy as follows.
\begin{equation}
    T_b = T_s - T_c = h(2d-1)n/l - 4un/b
\label{eq:benefit}
\end{equation}

Equation~\eqref{eq:benefit} shows the benefit is positively related to the number of hits and negatively related to the number of copies. Furthermore, for datasets with higher cardinality and dimensionality, the benefit is higher.

\section{Experimental study}
\label{paper:exp}
In this section, we present the empirically results of our proposed caching strategy HCST. We conducted the experiments on a workstation running Linux with two Xeon E5-2640v4 10 core CPUs and 256GB main memory. The number of threads is set to 20 by default to effectively utilize the resources of CPUs. The maximum number of items that cache can store is set to 5,000 by default, which is a relatively small size with little memory cost. We also try different cache sizes (i.e., 10K, 15K and 20K) in our experiments. The number of iterations between two consequent checkpoints is set to 20, which is an appropriate value as we will show in Section~\ref{paper:tune}. The cache replacement strategies we used in our experiments include HCST, LRU, EFU, LFU and LAT. To ensure fairness, we used the same data structure to implement all the cache strategies. The kernel functions we used in our experiments include the Gaussian kernel and the sigmoid kernel. We used ThunderSVM as our training library and the stopping criteria are the same for all the experiments. We used 14 public datasets from the LIBSVM website\footnote{\url{https://www.csie.ntu.edu.tw/~cjlin/libsvm/index.html}} and this link\footnote{\url{http://manikvarma.org/downloads/XC/XMLRepository.html}} with four different problems including binary, multi-label and multiclass classification, and regression. Among 14 datasets, 6 datasets (\emph{adult}, \emph{connect-4}, \emph{mnist}, \emph{webdata}, \emph{real-sim} and \emph{rcv1}) are used in every experiment while the other 8 datasets are used together with the previous datasets to more thoroughly evaluate the training time of SVMs with HCST. For \emph{mediamill} and \emph{rcv1s2}, the label dimensionality is 101. For \emph{amazon} and \emph{wikipedia}, we randomly choose 10 labels to perform the classification task. Table~\ref{tbl:dataset} gives the details of the datasets and parameters of kernel functions used in the experiments. The parameters are the same as the existing study~\cite{wen2014mascot,cotter2011gpu,tyree2014parallel} or selected by a grid search ($C$ ranges from 1 to 10 and $\gamma$ ranges from 0.001 to 0.5).

\begin{table}[!t]
\centering
\caption{datasets and kernel parameters}
\vspace{-5pt}
\label{tbl:dataset}
\resizebox{3.5in}{!}{%
\begin{tabular}{|c|c|c|c|c|c|c|c|} \hline
\multirow{2}{*}{dataset}	  & \multirow{2}{*}{task} & \multirow{2}{*}{cardinality}	& \multirow{2}{*}{dimension} & \multicolumn{2}{c|}{Gaussian} & \multicolumn{2}{c|}{sigmoid} \\ \cline{5-8} 
                 &&&  & $C$	& $\gamma$ & $C$	& $\gamma$	\\\hline
adult	  & \multirow{5}{*}{\begin{tabular}[c]{@{}c@{}}binary\\ classification\end{tabular}}       & 32,561		& 123		& 100	& 0.5	& 10 & 0.01	\\ \cline{1-1}\cline{3-8}
rcv1	  &    	& 20,242		& 47,236	& 100	& 0.125	&10 & 0.5\\\cline{1-1}\cline{3-8}
real-sim  &        & 72,309        & 20,958    & 4     & 0.5   &10 &0.5\\\cline{1-1}\cline{3-8}
webdata	  &        & 49,749		& 300		& 10	& 0.5	&10 &0.01	\\ \cline{1-1}\cline{3-8}
covtype   &        & 581,012       & 54        & 3     & 1 & 10 & 0.5 \\ \hline
connect-4 & \multirow{3}{*}{\begin{tabular}[c]{@{}c@{}}multiclass\\ classification\end{tabular}}   	& 67,557	    & 126		& 1	    & 0.3	& 10 & 0.001\\ \cline{1-1}\cline{3-8}
mnist	  &       & 60,000		& 780		& 10	& 0.125	&10 & 0.001\\\cline{1-1}\cline{3-8}
mnist8m   &      & 8,100,000     & 784       & 1000  & 0.006 & \backslashbox{}{} & \backslashbox{}{} \\\hline
mediamill & \multirow{4}{*}{\begin{tabular}[c]{@{}c@{}}multi-label\\ classification\end{tabular}}
     & 30,993 &12,914 &10 & 0.5 &10 &0.01 \\ \cline{1-1}\cline{3-8}
rcv1s2 & &3,000 & 101 & 10 & 0.5 & 10 & 0.5 \\ \cline{1-1}\cline{3-8}
amazon & &1,717,899 &337,067 &10 &0.5 &10 &0.5 \\ \cline{1-1}\cline{3-8}
wikipedia & &1,813,391 &2,381,304 &10 &0.5 &10 &0.5 \\ \hline
abalone & \multirow{2}{*}{\begin{tabular}[c]{@{}c@{}}regression\end{tabular}} &4,177 & 8 & 10 & 0.5 & 10 & 0.01 \\ \cline{1-1}\cline{3-8}
E2006-tfidf & & 16,087 & 150,360 & 10 & 0.5 & 10 & 0.01 \\ \hline
\end{tabular}
}
\vspace{-10pt}
\end{table}

\begin{figure}[!t]
\captionsetup[subfloat]{farskip=2pt,captionskip=1pt}
\centering
\subfloat[Hit ratio with the Gaussian kernel]{\includegraphics[width=1.75in]{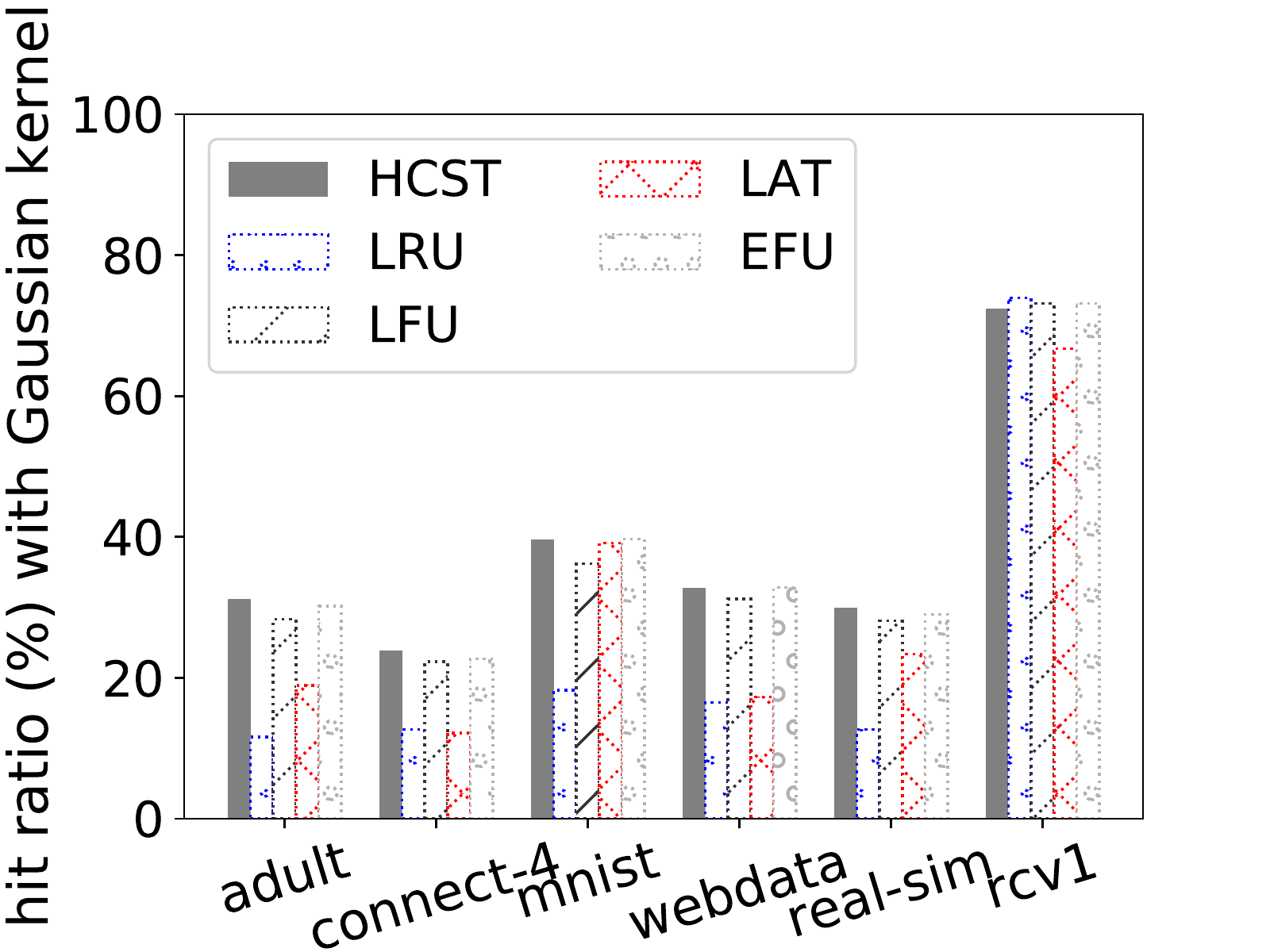}%
\label{fig:hitratiorbf}}
\subfloat[Hit ratio with the sigmoid kernel]{\includegraphics[width=1.75in]{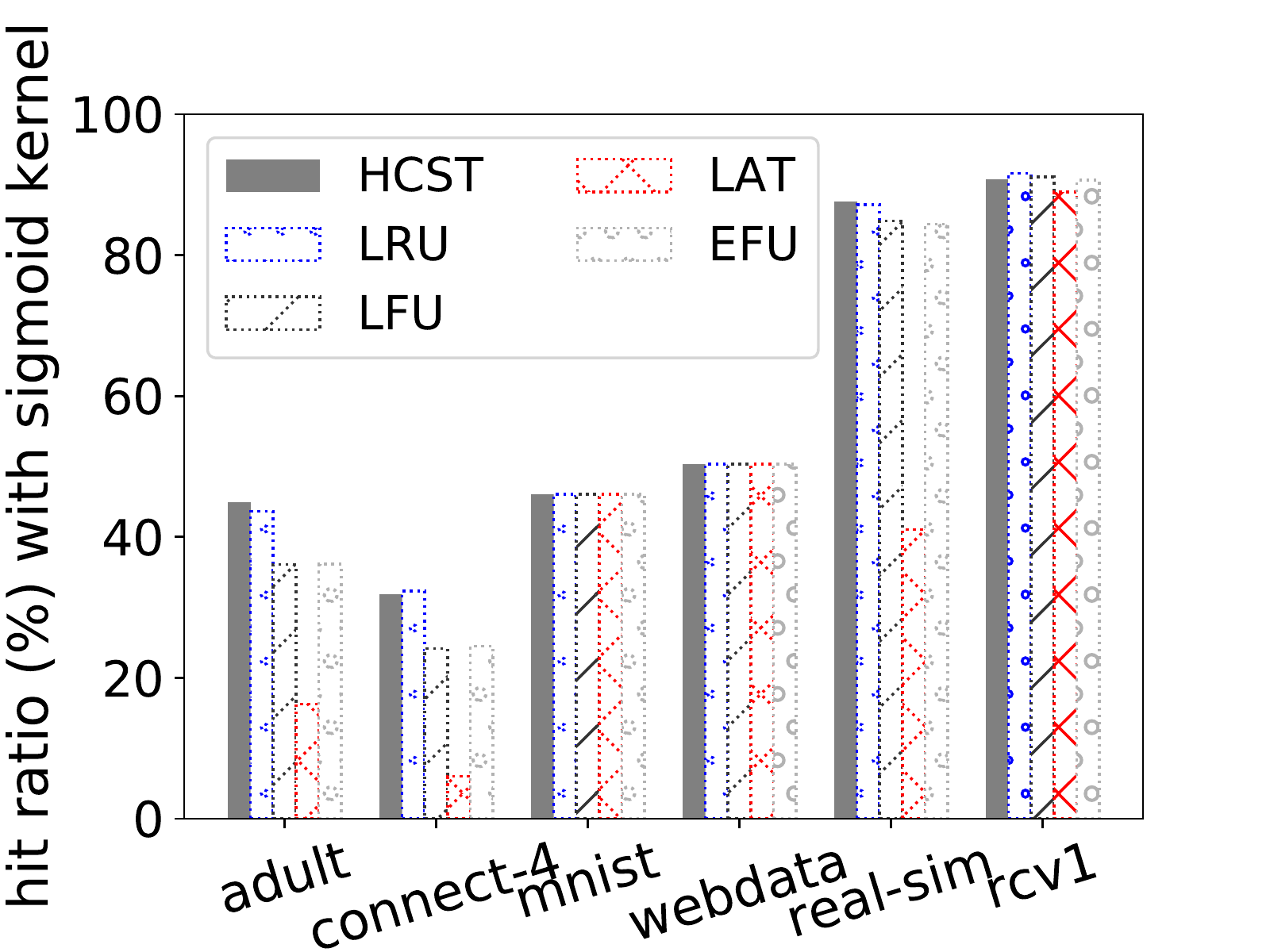}%
\label{fig:hitratiosigmoid}}
\caption{Hit ratio comparison}
\label{fig:hit}
\vspace{-15pt}
\end{figure}

\begin{figure}[!t]
\captionsetup[subfloat]{farskip=2pt,captionskip=1pt}
\centering
\subfloat[adult]{\includegraphics[width=1.75in]{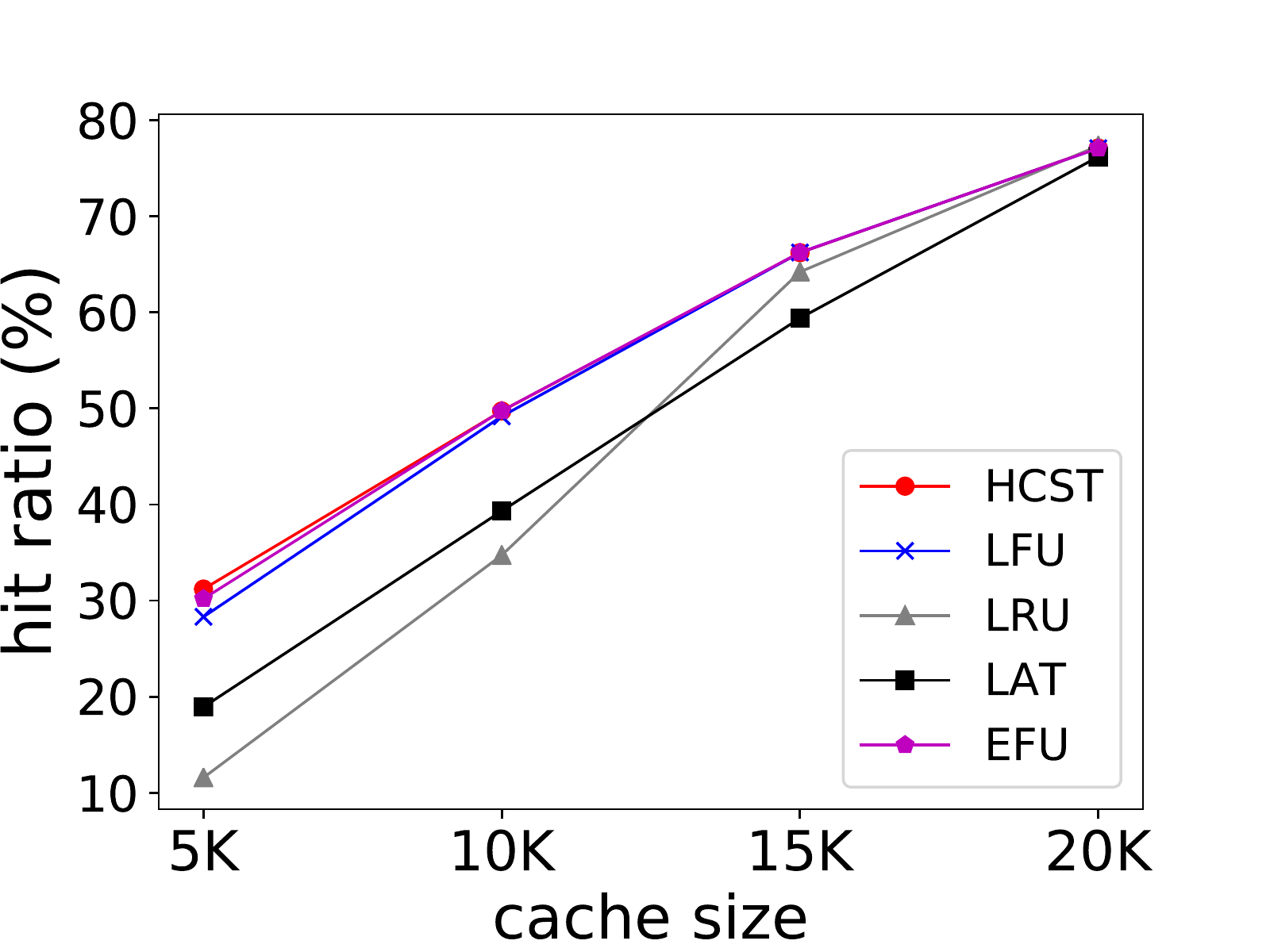}%
\label{fig:hitAdult}}
\subfloat[connect-4]{\includegraphics[width=1.75in]{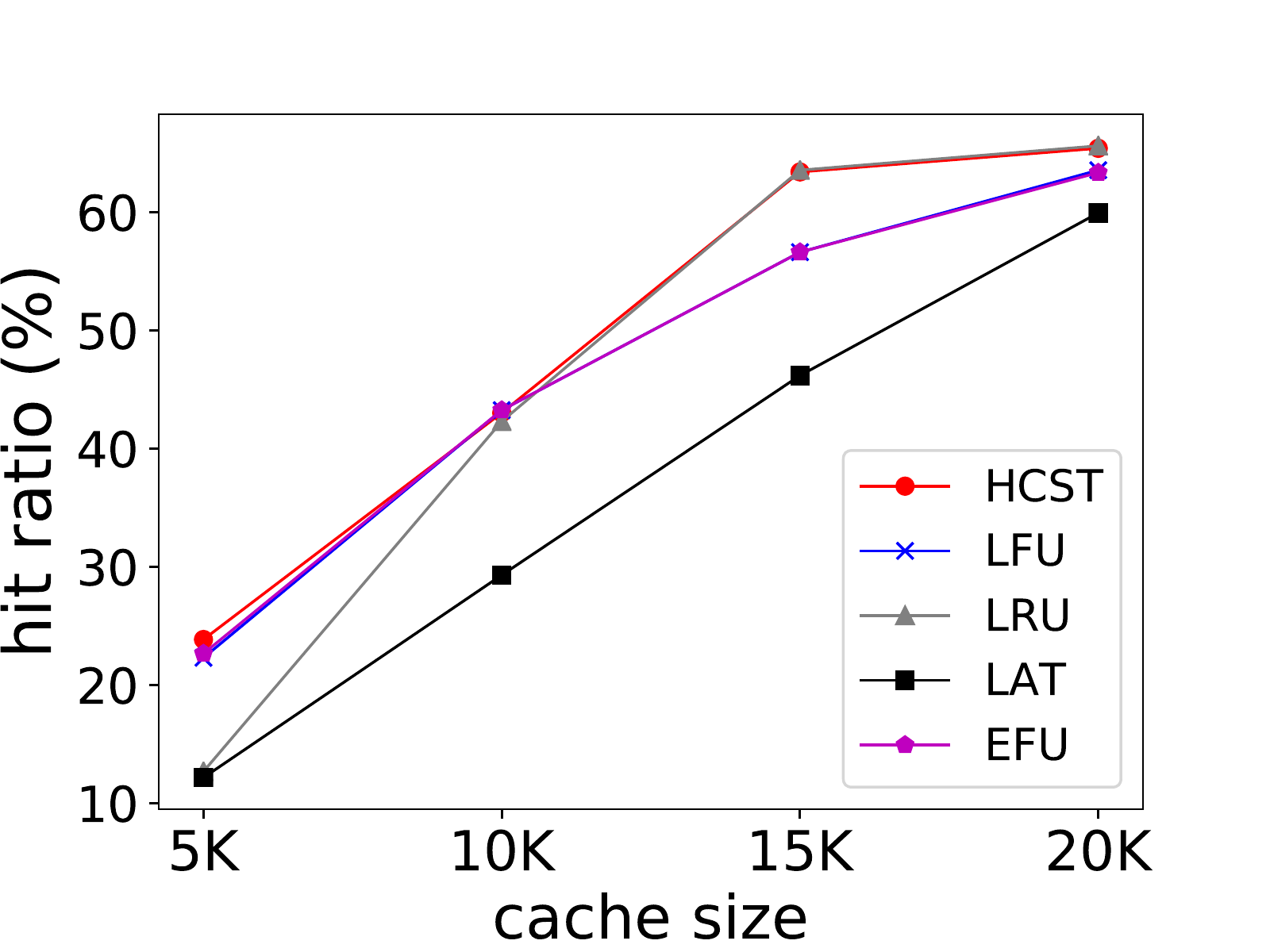}%
\label{fig:hitConnect}}

\subfloat[mnist]{\includegraphics[width=1.75in]{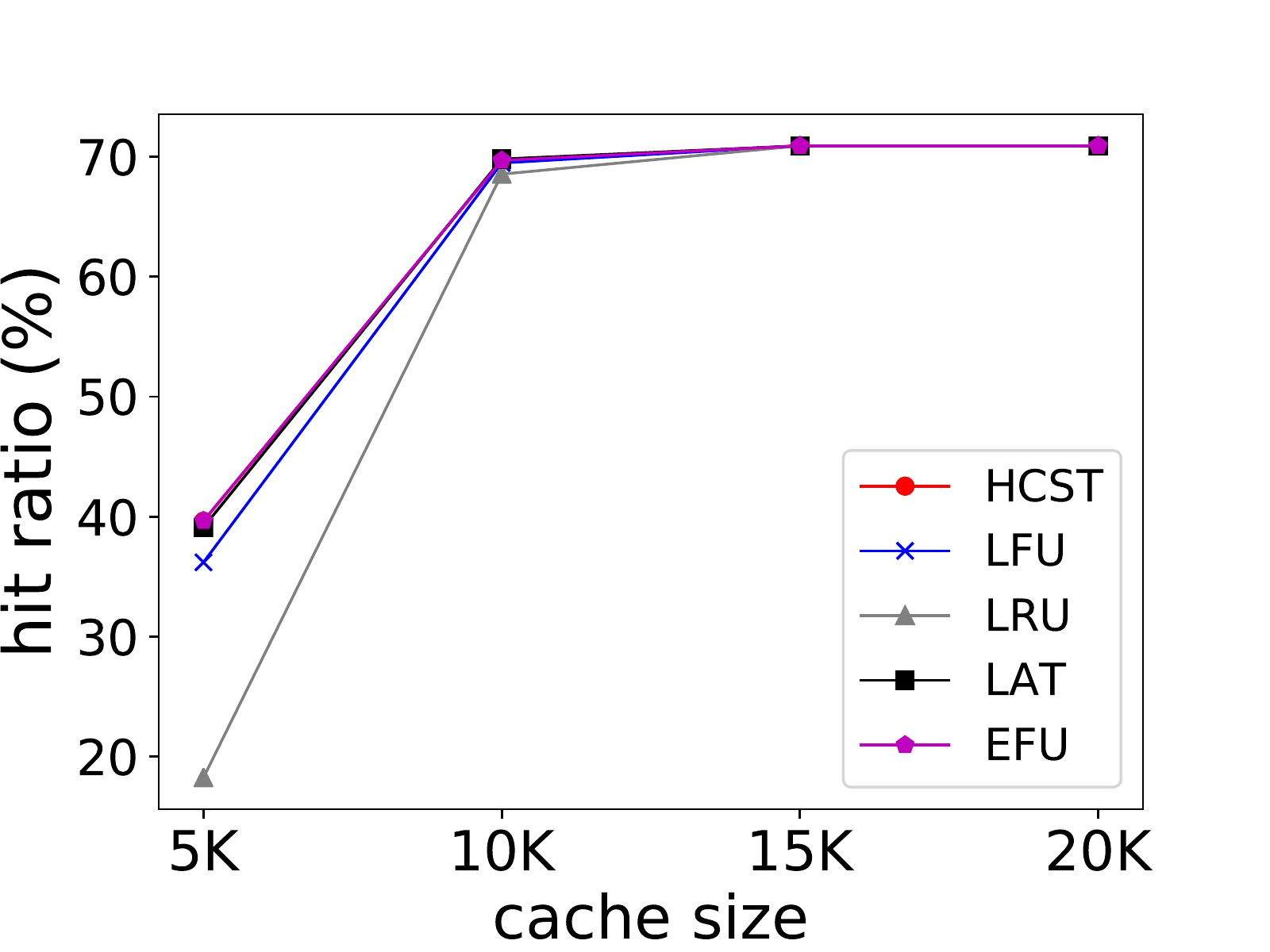}%
\label{fig:hitMnist}}
\subfloat[webdata]{\includegraphics[width=1.75in]{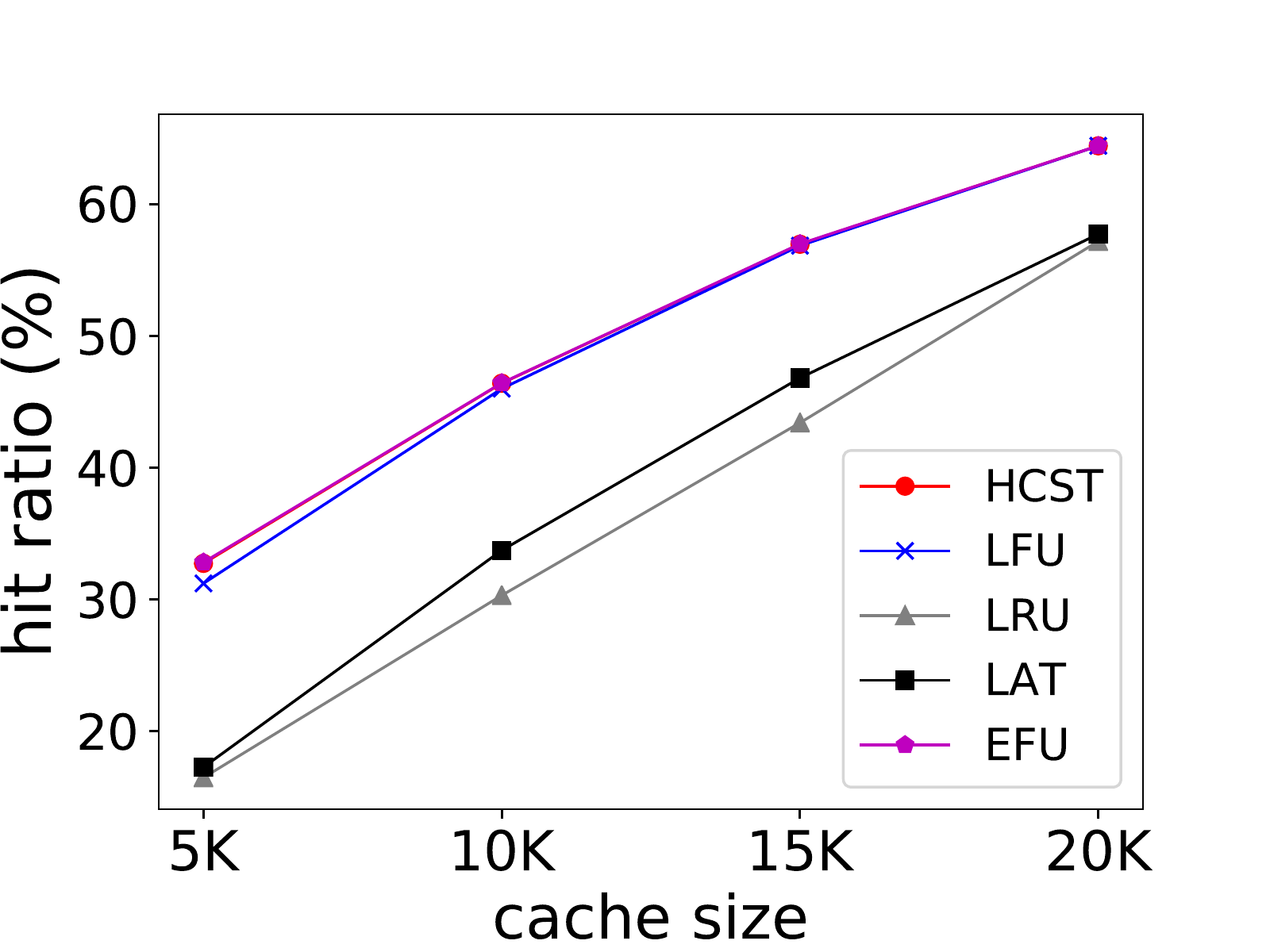}%
\label{fig:hitWeb}}

\subfloat[real-sim]{\includegraphics[width=1.75in]{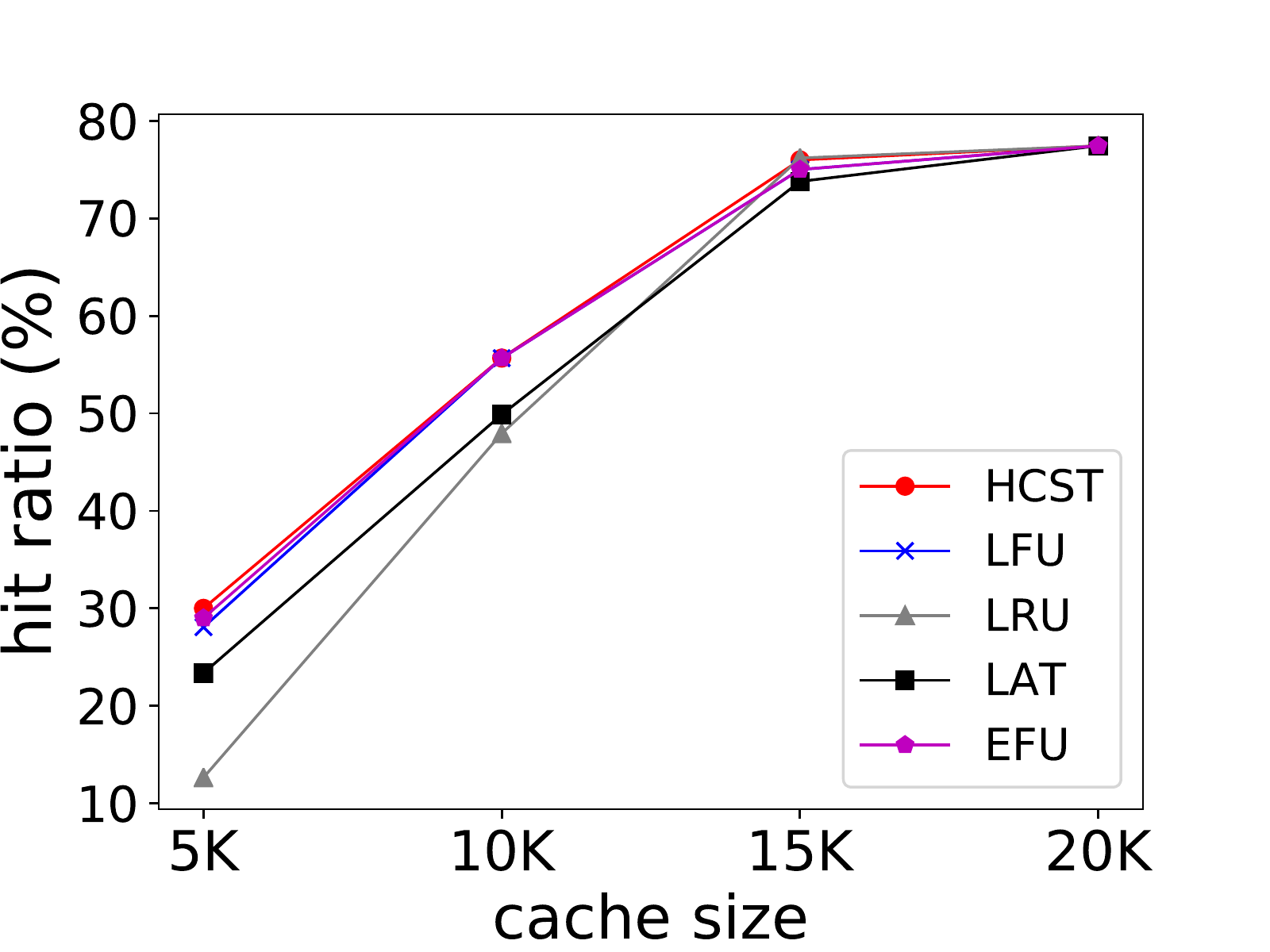}%
\label{fig:hitReal}}
\subfloat[rcv1]{\includegraphics[width=1.75in]{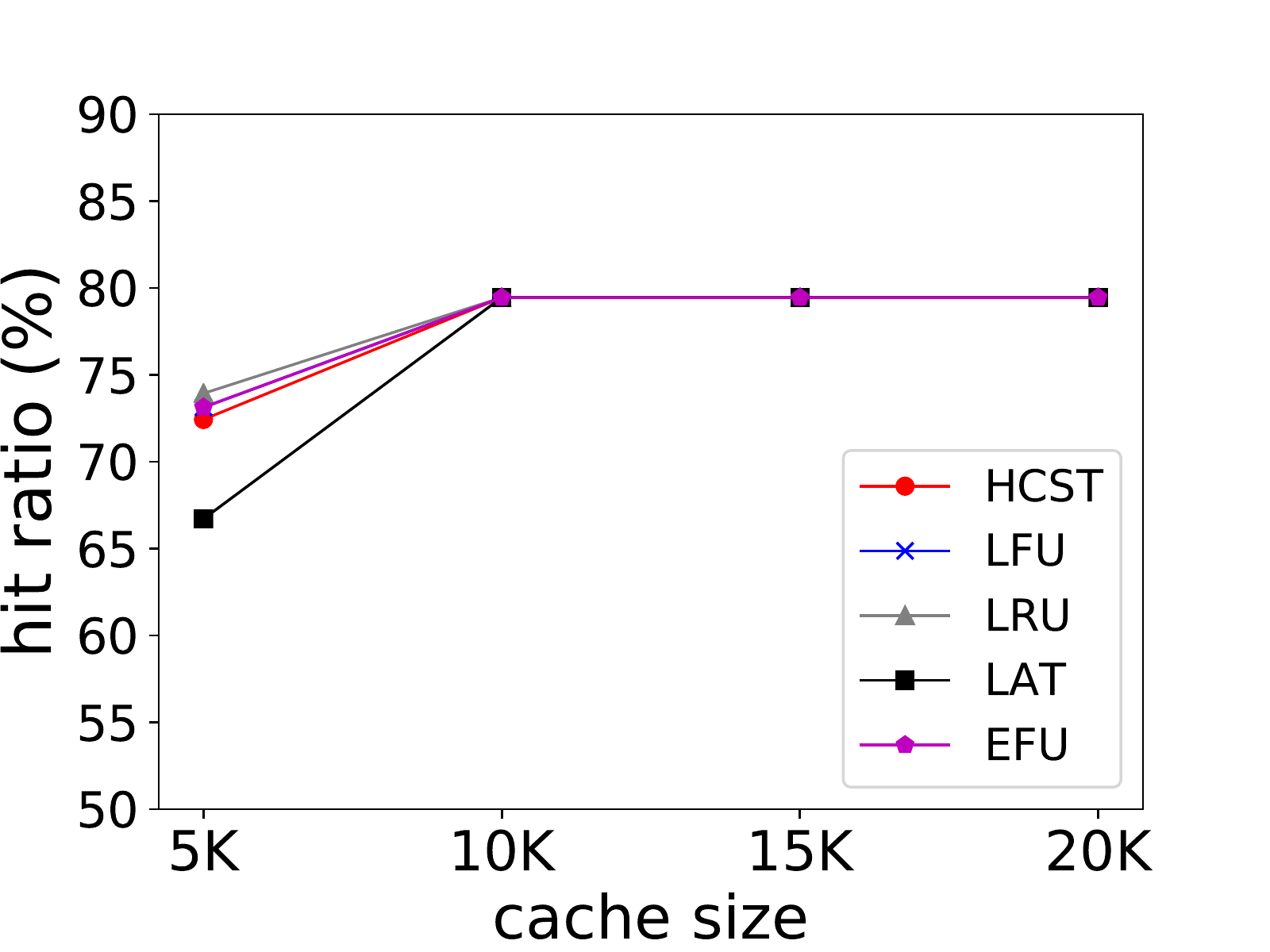}%
\label{fig:hitRcv1}}
\caption{Hit ratio with different cache sizes}
\label{fig:hitCacheSize}
\vspace{-10pt}
\end{figure}

\subsection{Hit ratio comparison}
\label{paper:hitratio}
\emph{Key finding 1: HCST can always achieve the highest hit ratio among all the caching strategies.}

Figure~\ref{fig:hit} shows the hit ratio of different caching strategies. We tried two different kernel functions: the Gaussian kernel and the sigmoid kernel. We first compare the four caching strategies except HCST. EFU and LFU have a relatively better performance when using the Gaussian kernel, while LRU outperforms the other strategies when using the sigmoid kernel. Furthermore, the hit ratio of EFU is higher than LFU on some datasets with the Gaussian kernel (e.g., \emph{adult} and \emph{mnist}). The LAT strategy does not show advantage on both two kernel functions. Except for HCST, none of the strategies is optimal across different datasets and different kernel functions. However, due to our specialized design, HCST is comparable to or even better than the best of them in all cases. HCST can adaptively achieve high hit ratios among different cases.

Figure~\ref{fig:hitCacheSize} shows the hit ratio of different caching strategies with different cache sizes using the Gaussian kernel, which is the most widely used kernel function in the SVM training. The HCST strategy always has good performance on different settings of cache size in the experiments. Furthermore, all five cache replacement strategies have a similar hit ratio when the cache size is big. This is because when the cache is large enough, the cache can store almost all the items. As a result, the accesses that yield cache misses are mainly first usages of the items, which are inevitable no matter what caching strategy is used. Note that the increase in memory size can not keep up with the speed of data growth. The memory size is always relatively small compared with the size of kernel values of huge datasets. The HCST strategy is superior to the other strategies when the memory for cache is limited.

\subsection{The overall training with HCST}
\emph{Key finding 2: HCST can always clearly reduce the SVM training time.}

To show the performance of the HCST strategy, we compare it with the cache strategies LRU, LFU, LAT and EFU on all the 14 datasets. In Table~\ref{tbl:trainTimeGaussian}, we show the elapsed training time of no cache and the relative values of the other caching strategies against no cache. Compared with HCST, the other strategies have a smaller improvement, and sometimes even have no benefit due to the cost of caching (e.g., \emph{adult}). The HCST strategy has a stable improvement and helps improve the SVM training without caching by at least 25\% in all 14 datasets. For \emph{rcv1}, the speedup is bigger than 4. For the multi-label datasets, HCST can reduce the training time by at least 30\%, which is a significant improvement.

\begin{table*}[!t]
\centering
\vspace{-5pt}
\caption{Comparison among HCST and the other existing caching strategies}
\label{tbl:trainTimeGaussian}
\resizebox{\textwidth}{!}{%
\begin{tabular}{|c|c|c|c|c|c|c|c|c|c|c|c|c|c|c|c|c|}
\hline
\multirow{3}{*}{dataset} &\multicolumn{8}{c|}{Gaussian kernel} &\multicolumn{8}{c|}{sigmoid kernel}\\ \cline{2-17} 
&\multicolumn{6}{c|}{elapsed time (sec) / relative value against no cache}  &\multicolumn{1}{c|}{speedup}  &\multicolumn{1}{c|}{training error}& \multicolumn{6}{c|}{elapsed time (sec) / relative value against no cache}  &\multicolumn{1}{c|}{speedup}  &\multicolumn{1}{c|}{training error}\\ \cline{2-7} \cline{10-15}
                         & no cache & HCST    & LRU & LFU  & LAT~\cite{wen2014mascot} &EFU
                         & \multicolumn{1}{c|}{of HCST}  & \multicolumn{1}{c|}{/RMSE}
                         & no cache & HCST    & LRU & LFU  & LAT &EFU 
                         & \multicolumn{1}{c|}{of HCST} & \multicolumn{1}{c|}{/RMSE}\\ \hline
adult                    &   24.65   &  -31.4\%      &+1.1\% & -9.5\% & -0.6\%  &-17.6\% & \textbf{1.46}            & 4.4\% &6.86 &-24.2\% &-13.3\% &-2.6\% &+5.0\% &-6.9\% &\textbf{1.32} & 15.2\% \\ \hline
connect-4                &  91.61     &-26.0\%       & +2.1\% & -11.0\% & +3.1\% &-18.8\% & \textbf{1.35}            & 4.39\% &58.34 &-20.6\% &-18.5\% &-8.2\% &-2.7\% &-10.9\%& \textbf{1.26}& 24.42\%\\ \hline
mnist                    &  346.12   &-36.2\%   & -9.1\% &-26.0\%  & -29.6\% &-33.5\% & \textbf{1.57}            & 0\% &40.02 &-28.8\% &-26.3\% &-23.5\% &-25.1\% &-28.3\% &\textbf{1.40} &5.57\%\\ \hline
webdata                     & 38.21   &-23.9\%    & -0.4\% &-11.2\%  & 0\% & -17.7\% &\textbf{1.31}  & 0.54\%  &2.89 &-33.9\% &-24.6\% &-23.9\% &-23.5\% &-30.1\% &\textbf{1.51} & 1.18\%\\ \hline
real-sim                 &  75.93    &  -34.4\%    & -11.2\% & -28.1\%  & -21.8\% &-33.9\% & \textbf{1.52}   & 0.27\%  &239.13 &-79.0\% &-76.7\% &-72.6\% &-42.8\% &-75.7\% &\textbf{4.76} & 0.84\%\\ \hline
rcv1                  & 27.66     & -77.3\% & -78.9\%  & -78.3\% & -73.9\% &-77.2\%& \textbf{4.41}  & 0.11\% &57.87 &-89.6\% &-89.4\% &-88.9\% &-88.5\% &-89.4\% &\textbf{9.59} & 0.29\%\\ \hline
mnist8m                  & 219142  & -45.7\%    & -34.2\% &-39.9\% & -7.1\% &-40.5\% & \textbf{1.84}   & 0\%     &\multicolumn{6}{c|}{$> 7$ days}& \backslashbox{}{} & \backslashbox{}{}   \\ \hline
covtype                  & 2819        &  -23.7\%    & -14.5\% & -19.8\%  & -15.4\%  &-20.7\% & \textbf{1.31}   & 0\%    &2164 &-20.2\% &-10.9\% &-9.9\% &-10.9\% &-20.6\% &\textbf{1.25} & 51.24\%   \\ \hline
mediamill & 448.54 & -33.9\% & -22.9\% & -20.9\% & -16.1\% & -21.5\% & \textbf{2.52} & 2.81\% &468.48 &-44.7\%  & -30.3\% & -27.4\% & -18.2\% & -31.0\% & \textbf{1.81} & 3.13\% \\ \hline
rcv1s2 & 252.48 &-98.3\%  & -62.0\% & -62.2\% & -62.2\% & -61.9\% & \textbf{57.4} &1.97\% &243.74 &-98.5\%    & -61.3\% & -61.4\% & -61.7\% & -62.2\% & \textbf{65.0} & 1.98\%\\ \hline
amazon & 22317 & -33.5\% &-9.0\%  &-21.5\%  &-13.6\%  &-29.5\%  &\textbf{1.50}  &0\%  & 1770 &-42.0\%  &-21.7\% &-21.8\%  &-21.9\% &-24.5\%  &\textbf{1.73} &0\%  \\ \hline
wikipedia & 75384  & -49.6\% &-15.6\%  &-27.8\%  &-21.6\%  &-35.0\%  & \textbf{1.98} &0\% &10913  &-52.2\% &-30.5\%  &-28.6\% &-25.5\%  &-27.3\% &\textbf{2.09}&5.93\%  \\ \hline
abalone & 0.53 & -30.2\% & -17.0\% & -18.9\% & -20.8\% & -22.6\% & \textbf{1.43} & 4.54  & 0.41 & -22.0\% & -12.2\% & -17.1\% & -17.1\% & -19.5\% & \textbf{1.28} &6.46 \\ \hline
E2006-tfidf & 224.43 & -48.0\% & -38.4\% & -33.1\% & -36.1\% & -34.0\% & \textbf{1.92} & 0.12 & 219.54 & -26.6\% & -23.0\% & -13.8\% & -18.5\% & -19.2\% & \textbf{1.36} & 0.50 \\ \hline
\end{tabular}%
}
\vspace{-15pt}
\end{table*}

Table~\ref{tbl:partTime} shows the calculation time of kernel values and the cost of caching when using different caching strategies with the Gaussian kernel. Compared with LFU, LRU, LAT and EFU, HCST always has lower calculation time of kernel values and lower cost of caching. As we have shown in Figure~\ref{fig:hitratiorbf}, the HCST strategy mostly has a higher hit ratio than the other strategies, especially LRU and LAT, which explains the lower calculation time of kernel values. As we have discussed in Section~\ref{paper:paraReplace}, the replacement of HCST can be performed in parallel, which indicates the lower cost of updating cache.

\begin{table}[!t]
\centering
\caption{Calculation time of kernel values and cost of caching (sec)}
\label{tbl:partTime}
\vspace{-5pt}
\resizebox{3.5in}{!}{%
\begin{tabular}{|c|c|c|c|c|c|c|c|c|c|c|c|}
\hline
\multirow{3}{*}{dataset} & \multicolumn{6}{c|}{calculation time of kernel values} & \multicolumn{5}{c|}{overhead of cache replacement} \\ \cline{2-12} 
  & \multirow{2}{*}{no cache} &\multicolumn{5}{c|}{relative value against no cache} & \multirow{2}{*}{HCST} & \multicolumn{4}{c|}{relative value against HCST} \\ \cline{3-7} \cline{9-12}
                         &     & HCST     & LRU    & LFU  & LAT  &EFU   &        & LRU  & LFU   & LAT &EFU    \\ \hline
adult                      & 18.32       & -42.6\%  & -9.5\%   &-26.9\% &-14.7\% &-35.3\%   & 0.19    &+1405\% & +1463\%  &+1363\% &+1047\%    \\ \hline
connect-4                & 75.76       & -33.6\%    & -9.4\% &-26.9\%  &-11.0\% &-26.6\% & 0.84   &+910\% & +952\%  &+988\% &+238\%  \\ \hline
mnist                    & 315.7      & -38.7\%    & -15.8\%  &-32.54\%  &-37.7\% &-38.1\% & 1.85  &+773\% & +823\%  &+544\% &+541\%  \\ \hline
webdata                 & 29.63       & -33.5\%    &-16.8\%   &-30.4\% &-17.4\% &-32.7\%   & 0.20   &+2170\%   &+2220\%  &+2365\% &+1010\%  \\ \hline
real-sim                 &  67.1      & -40.7\%   &-20.6\% &-39.2\% &-30.3\%  &-41.9\% & 0.33    &+1321\% & +1227\% &+1030\%  &+315\%  \\ \hline
rcv1                  & 26.36       & -82.1\%     &-83.7\% &-83.2\% &-78.8\%  &-77.4\% & 0.06     &+66.7\% & +133\% &+167\% &0\%   \\ \hline
mnist8m                  & 213595 & -48.2\%       &-37.0\% &-42.5\% &-8.3\% &-43.0\%  & 2380    &+27.0\%  & +14.4\% &+92.7\% &+5.9\%\\ \hline
covtype                 &2760 &-43.4\% &-36.3\% &-40.2\%  &-37.7\% &-41.2\%         &319 &+25.7\% &+12.2\% &+28.5\% &+1.9\%\\ \hline
mediamill             & 392.36 &-41.7\%  & -33.5\% &-30.4\% &-21.1\% &-30.5\% & 5.32 &+130\% &+168\% &+338.2\% &0\% \\ \hline
rcv1s2               & 249.86 &-99.5\% &-62.9\% &-63.0\% &-63.0\% &-62.8\% & 0.31 &+9.7\% &0\% &+3.2\% &0\% \\ \hline
amazon              &21371 &-35.8\% &-11.9\% &-24.5\% &-16.5\% &-30.9\% &29 &+1697\% &+1486\% &+1603\% &-13.8\% \\ \hline
wikipedia          &74820  &-50.3\% &-16.1\% &-28.3\% &-22.2\% &-35.3\% &5.9 &+5137\% &+4442\% &+5239\% &+290\% \\ \hline
abalone              & 0.37 &-45.9\% &-40.5\% &-43.2\% &-43.2\% &-40.5\% &0.02 &+200\% &+200\% &+150\% &0\%\\ \hline
E2006-tfidf         & 222.91 &-48.6\% &-38.7\%  &-33.6\%  &-36.5\% &-34.2\% &0.15 &+227\% &+300\% &+267\% &+120\%\\ \hline
\end{tabular}
}
\vspace{-10pt}
\end{table}

Moreover, we can estimate the speedup of HCST on a GPU. Specifically, the training time with HCST on a GPU can be approximately computed by the hit ratio, which is independent of the hardware. Suppose the hit ratio of HCST is $h$ and the calculation time of kernel values without cache is $t_k$. Then we can estimate the calculation time of kernel values with HCST as $ht_k$ and ignore the overhead of caching which is relatively small. We have conducted experiments with a Pascal P100 GPU of 12GB memory. Table~\ref{tbl:gpu_est} shows the exact training time without cache and the estimated training time with HCST. HCST can still reduce the training time by at least 10\%.

\begin{table}[!t]
\centering
\caption{The estimated training time on a GPU with HCST (sec)}
\label{tbl:gpu_est}
\vspace{-5pt}
\begin{tabular}{|c|c|c|c|c|c|}
\hline
\multirow{3}{*}{datasets} & \multicolumn{2}{c|}{calculation time of} & \multicolumn{2}{c|}{elapsed time of} & \multirow{3}{*}{\begin{tabular}[c]{@{}c@{}}speedup \\ of HCST\end{tabular}}  \\ 
 & \multicolumn{2}{c|}{kernel values (sec)} & \multicolumn{2}{c|}{the training (sec)} &  \\ \cline{2-5}
 & no cache & HCST & no cache & HCST & \\ \hline
adult & 750 & 516 & 2749 & 2515 & 1.09 \\ \hline
connect-4 & 2320 & 1766 & 6550 & 5996 & 1.09 \\ \hline
mnist & 15122 & 9134 & 33626 & 27638 & 1.22 \\ \hline
webdata & 1013 & 682 & 3421 & 3090 & 1.11 \\ \hline
real-sim & 2358 & 1651 & 4105 & 3398 & 1.21 \\ \hline
rcv1 & 795 & 219 & 1555 & 979 & 1.59 \\ \hline
\end{tabular}%
\vspace{-15pt}
\end{table}

\subsection{Effect of factors}
\label{sec:factor}

\begin{figure}[!t]
\captionsetup[subfloat]{farskip=2pt,captionskip=1pt}
\centering
\subfloat[Gaussian kernel]{\includegraphics[width=1.75in]{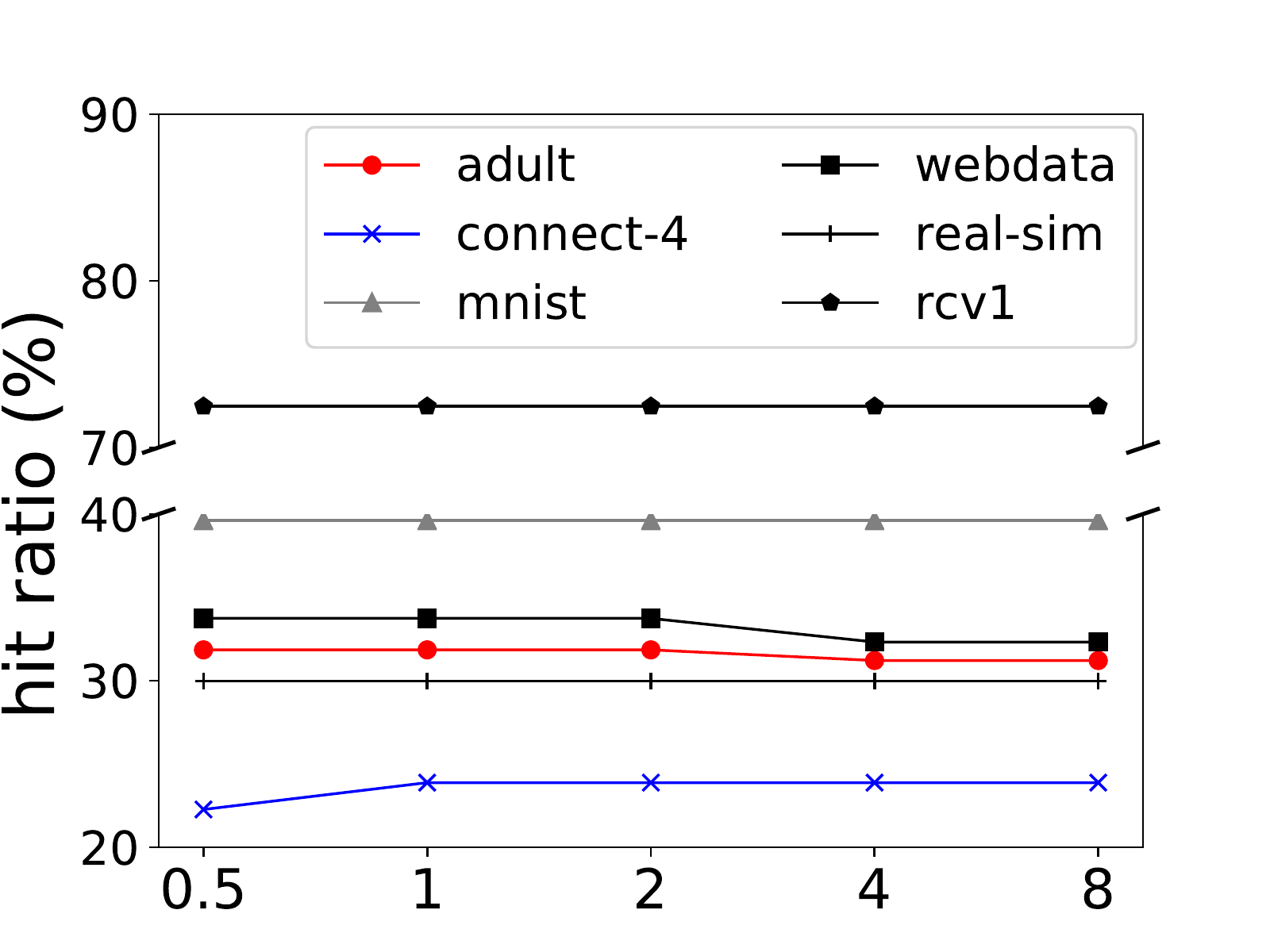}%
}
\subfloat[sigmoid kernel]{\includegraphics[width=1.75in]{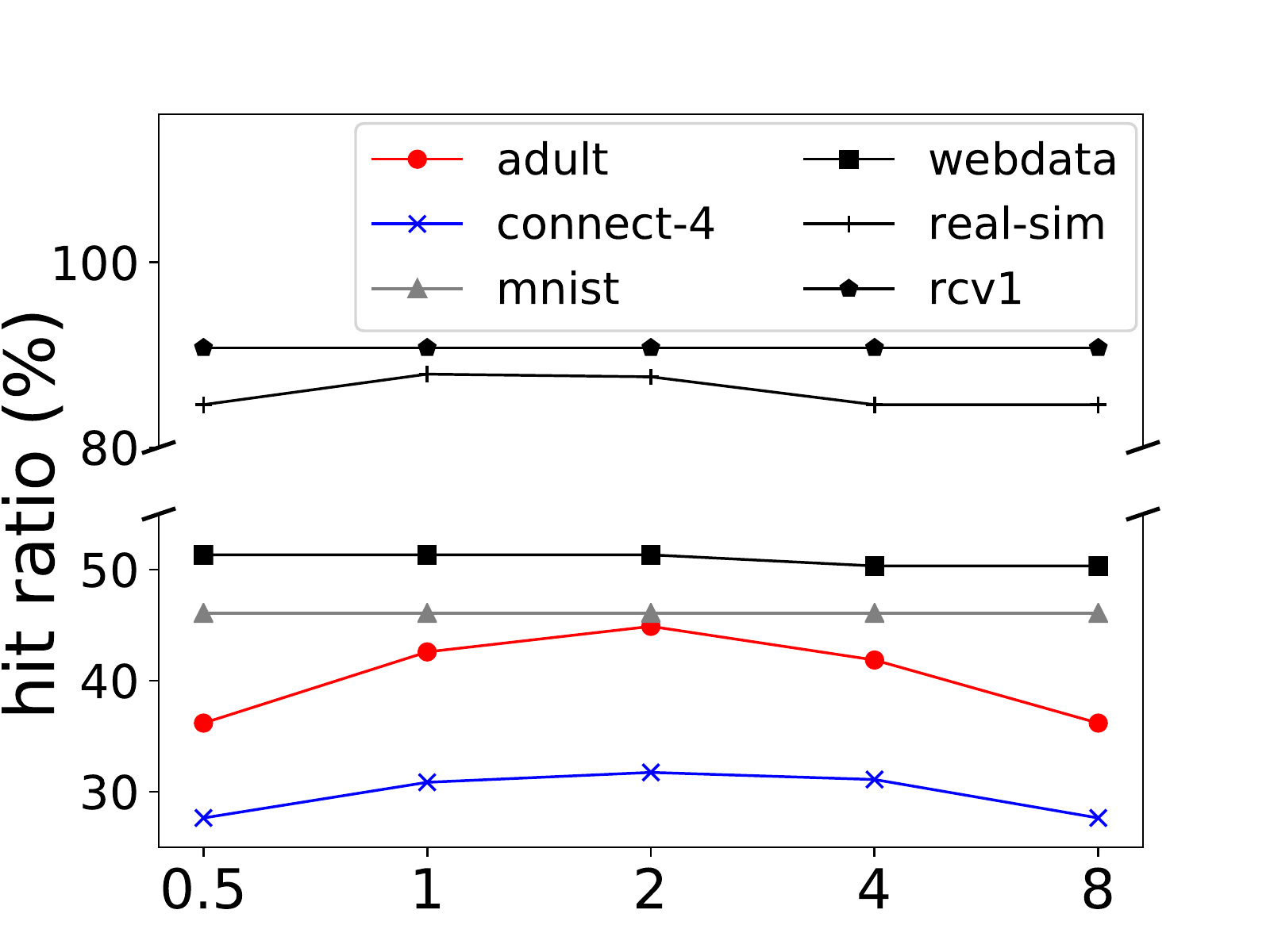}%
}
\caption{Hit ratio with different $\lambda$}
\label{fig:tuning}
\vspace{-15pt}
\end{figure}

\begin{figure}[!t]
\captionsetup[subfloat]{farskip=2pt,captionskip=1pt}
\centering
\subfloat[Gaussian kernel]{\includegraphics[width=1.75in]{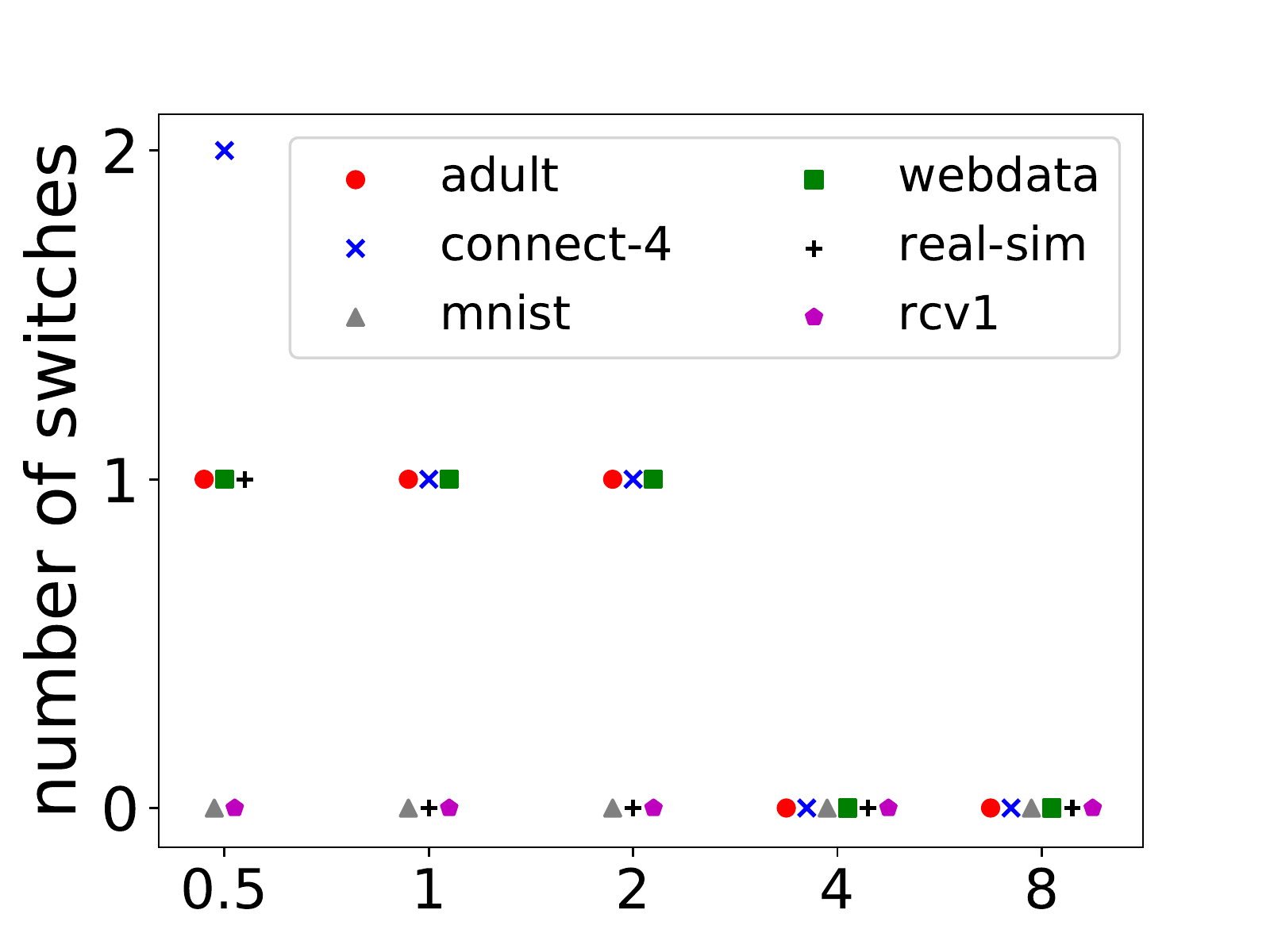}%
}
\subfloat[sigmoid kernel]{\includegraphics[width=1.75in]{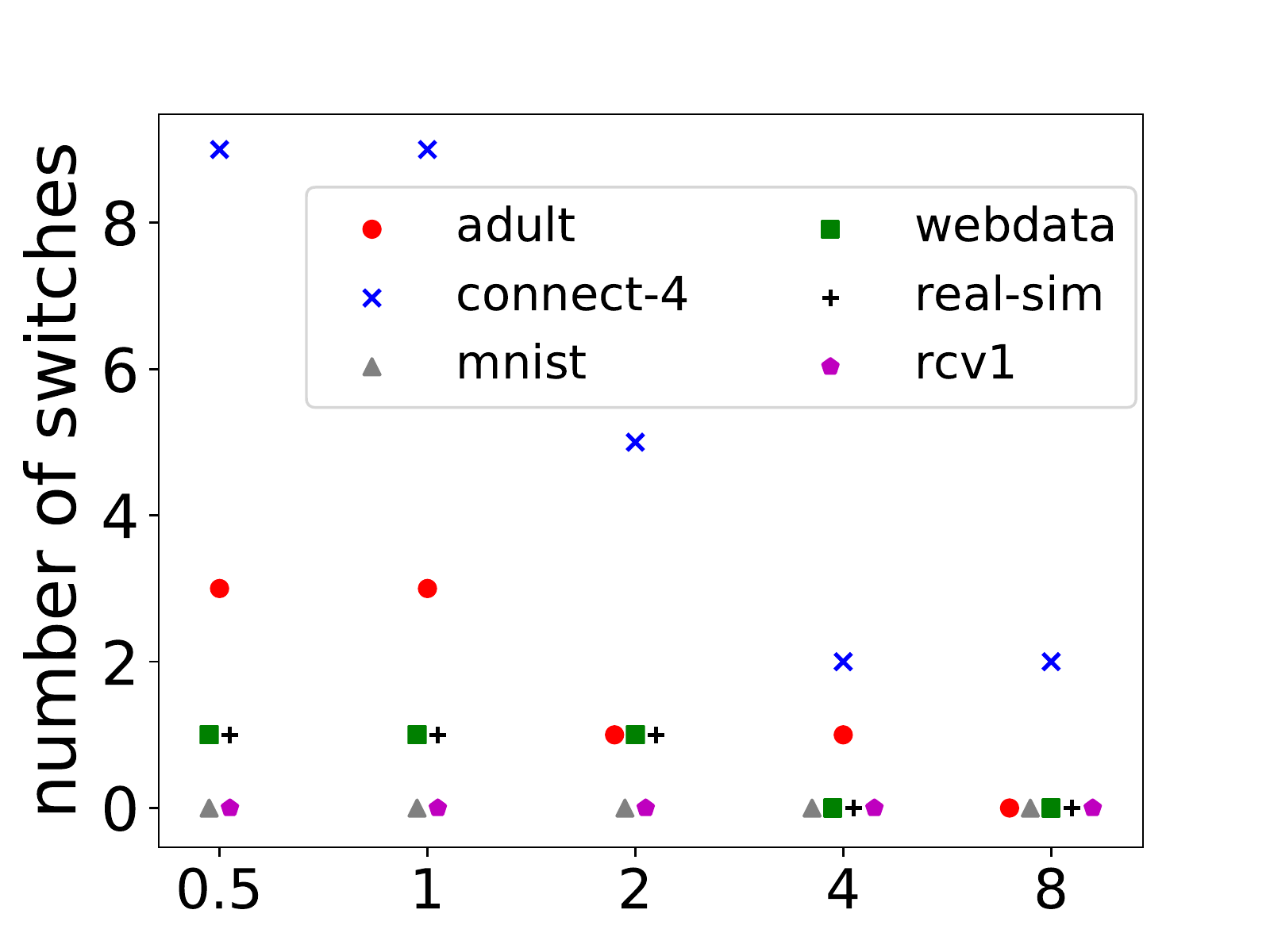}%
}
\caption{The number of switches with different $\lambda$}
\label{fig:nswitch}
\vspace{-15pt}
\end{figure}

\subsubsection{The impact of the length between two checkpoints}
\label{paper:tune}

\emph{By setting the number of iterations between two consequent checkpoints to $2s/q$, HCST can achieve the best performance.}

The setting of number of iterations between two consequent checkpoints should be appropriate. Suppose the number of iterations between two consequent checkpoints is $N_c$ and the cache size is $s$. Since ThunderSVM chooses $q$ items in each iteration, there are up to $qN_c$ replacement operations between two consequent checkpoints. We set the parameter $\lambda = qN_c / s$ and try different $\lambda$. The results are shown in Figure~\ref{fig:tuning}. Here we only show the experiments on six representative datasets, while the behavior of the other datasets is similar. From the results, we can observe that the hit ratio does not change clearly on \emph{mnist} and \emph{rcv1} with different $\lambda$. Also, we can find that the hit ratio changes more dramatically in the sigmoid kernel than the Gaussian kernel on \emph{adult} and \emph{connect-4}. To explain these cases, we have measured the number of switches of the training on different datasets, as shown in Figure~\ref{fig:nswitch}. For \emph{mnist} and \emph{rcv1}, the number of switches is always zero, which means EFU is always the appropriate strategy for the training. Then, the hit ratio does not change no matter the value of $\lambda$. For \emph{adult} and \emph{connect-4}, compared with the Gaussian kernel, the range of the number of switches among different $\lambda$ is bigger in the sigmoid kernel. Thus, the hit ratio also change more clearly in these cases. Generally, HCST has a well hit ratio on all cases when setting $\lambda = 2$, where $N_c = 2s/q$. Since $s=5000$ and $q=512$ in our experimental setting, we set the number of iterations between two consequent checkpoints to 20 (i.e., $2*5000/512$) by default.

\begin{figure}[!t]
\captionsetup[subfloat]{farskip=2pt,captionskip=1pt}
\centering
\subfloat[cost]{\includegraphics[width=1.75in]{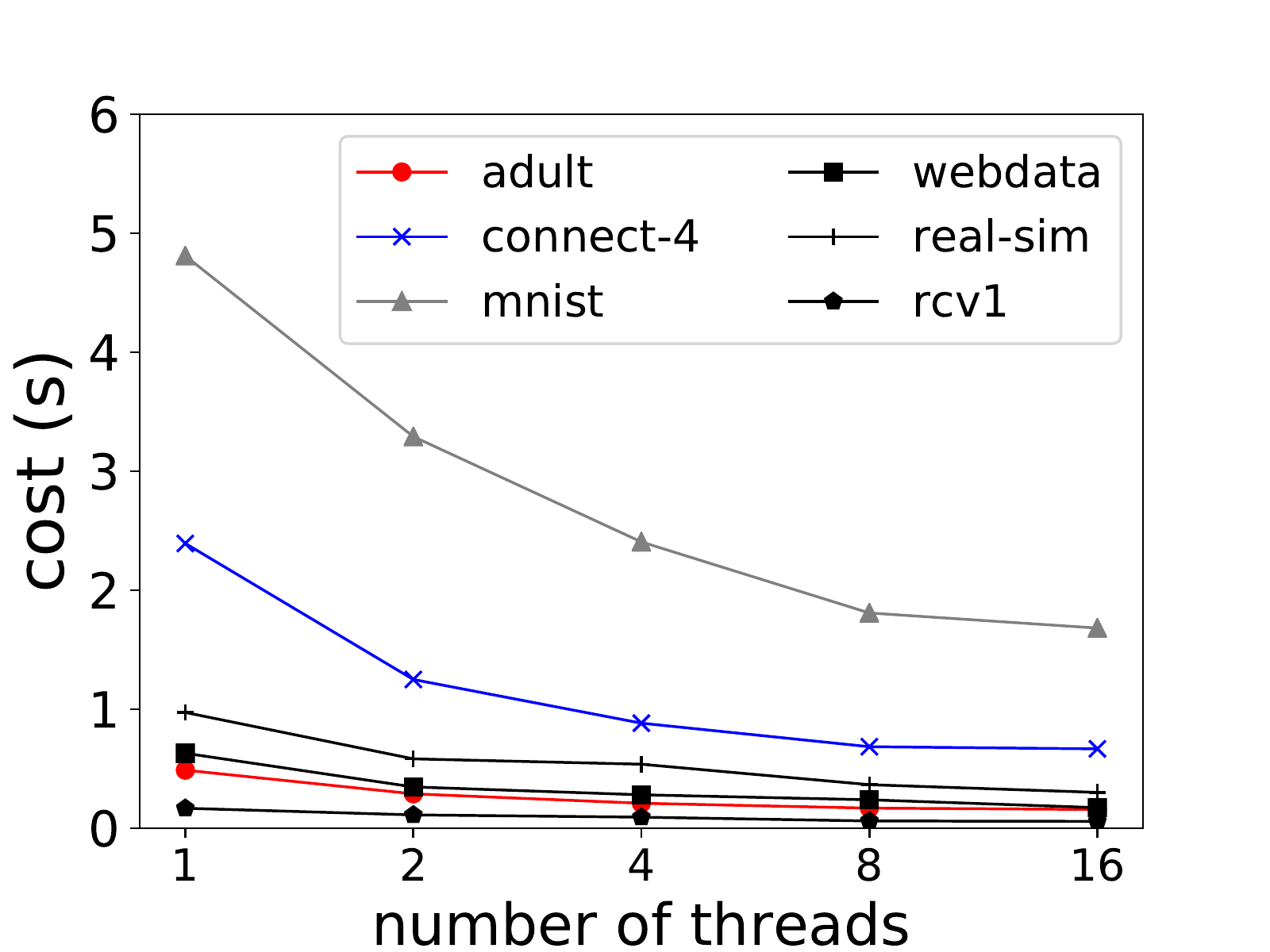}%
\label{fig:costthreads}}
\subfloat[hit ratio]{\includegraphics[width=1.75in]{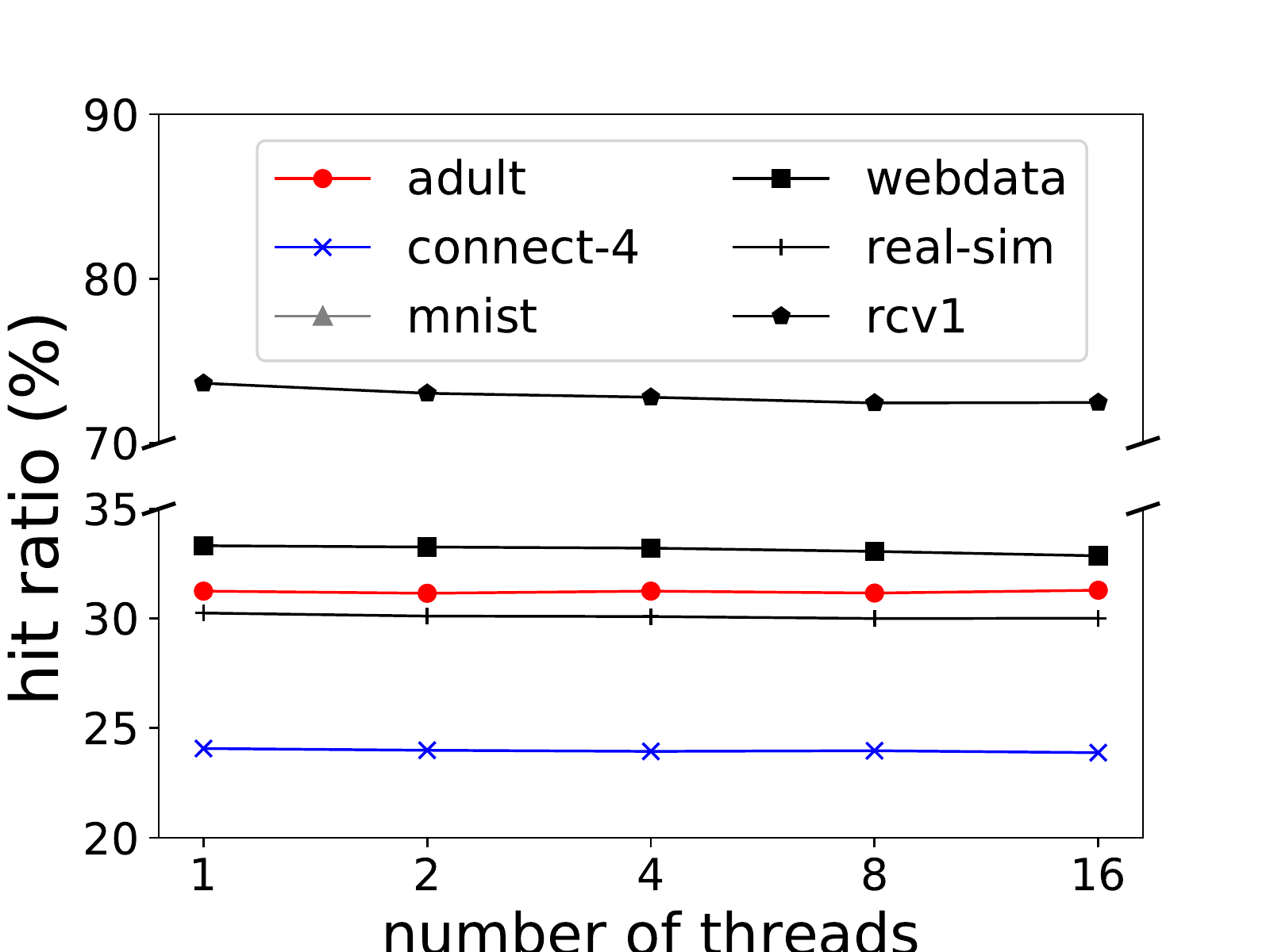}%
\label{fig:hitThreads}}
\caption{The parallel HCST strategy}
\label{fig:parallelRep}
\vspace{-15pt}
\end{figure}

\subsubsection{The impact of the number of threads on HCST}
\emph{The cost of HCST can be reduced significantly by using multiple threads while retaining the same hit ratio.}

Figure~\ref{fig:costthreads} shows the cost of the HCST caching strategy with different numbers of threads. Here, the ``cost'' of HCST means the extra computation results from the caching strategy, which includes copying the items to the cache and deciding whether the items are in the cache. As we can see from the results, the cost is significantly reduced by using multiple threads. When we use two threads, the cost can be reduced by more than 40\% compared with the sequential version of HCST. Parallel HCST can improve the bandwidth usage and can save notable amount of time on copying items to the cache.

Next, we inspect the effect of multi-threading on the hit ratio. Figure~\ref{fig:hitThreads} shows the hit ratio of HCST with different numbers of threads. The hit ratio is almost unchanged for all the datasets. This confirms that the parallel implementation of HCST has little impact on the hit ratio.

\subsubsection{The impact of different computational environments}
\emph{The HCST strategy is portable across different environments.}

To show the effect of HCST on different computational environments, we conducted the experiments on a supercomputer in National Supercomputing Centre Singapore. The server of the supercomputer has four Xeon E7-4830v3 12 core CPUs and 1TB memory. The number of threads is set to 48 and the cache size is set to 5000. The results are shown in Table~\ref{tbl:sctime}. We can still achieve over 1.2 times speedup by using HCST. Compared with the other strategies, HCST has a superior performance. The effectiveness of our caching strategy is stable regardless of the computational environments.

\begin{table}[!t]
\centering
\caption{Training time on a server of NSCC supercomputer (sec)}
\label{tbl:sctime}
\vspace{-5pt}
\resizebox{3.5in}{!}{%
\begin{tabular}{|c|c|c|c|c|c|c|c|}
\hline
    \multirow{2}{*}{dataset} &\multicolumn{6}{c|}{elapsed time (sec) / relative value against no cache} &speedup  \\ \cline{2-7}
   & no cache & HCST &LRU & LFU & LAT  &EFU  &of HCST \\ \hline
adult     & 26.74    & -22.7\% & +0.9\%  & -11.0\% & -2.1\%   & -20.1\% & 1.29            \\ \hline
connect-4 & 70.73    & -14.78\%  & +11.4\% & +2.2\% & +10.9\% & -7.9\%  & 1.17            \\ \hline
mnist     & 244.56   & -23.7\% & -1.1\%   & -11.1\% & -11.3\%  & -21.0\% & 1.31            \\ \hline
webdata   & 41.89    & -27.0\% & -1.5\%   & -12.8\% & +0.6\%  & -24.5\% & 1.37            \\ \hline
real-sim  & 60.89    & -38.3\% & -16.9\%  & -31.0\% & -27.4\%  & -35.8\% & 1.62            \\ \hline
rcv1      & 32.66    & -81.8\% & -76.0\%  & -77.9\% & -75.5\%  & -80.0\% & 5.48            \\ \hline
\end{tabular}
}
\vspace{-15pt}
\end{table}

\subsubsection{The impact of the two-level reuses in multi-output tasks}
\label{paper:solver-reuse}
\emph{The solver-level reuse of the kernel values can reduce the training time significantly.}

\del{To show the impact of the two-level reuses of the kernel values,}Table~\ref{tbl:two-level-reuses} shows the training time of no cache and the relative time with different kinds of reuses against no cache, where IR denotes the iteration-level reuse and SR denotes the solver-level reuse. On the basis of the original HCST (i.e., only IR), the SR technique can usually further reduce the training time by more than 10\%. For \emph{rcv1s2}, our approach can even reduce the time by more than 95\%, which is a significant improvement. The results show that our optimization for the multi-output learning tasks is quite effective.

\begin{table}[]
\centering
\caption{The impact of the two-level reuses}
\label{tbl:two-level-reuses}
\vspace{-5pt}
\resizebox{3.5in}{!}{%
\begin{tabular}{|c|c|c|c|c|c|c|}
\hline
\multirow{2}{*}{datasets} & \multicolumn{3}{c|}{Gaussian kernel} & \multicolumn{3}{c|}{sigmoid kernel} \\ \cline{2-7} 
 & no cache & only IR & IR and SR & no cache  & only IR & IR and SR \\ \hline
mediamill & 448.54 & -23.0\% & -33.9\% & 468.48 & -27.7\% & -44.7\% \\ \hline
rcv1s2 & 252.48 & -61.9\% & -98.3\% & 243.74 & -61.5\% & -98.5\% \\ \hline
amazon-3M & 22317 & -29.1\% & -33.5\% & 1770 & -27.6\% & -42.0\% \\ \hline
wiki-500k & 75384 & -35.6\% & -49.6\% & 10913 & -28.6\% & -52.2\% \\ \hline
\end{tabular}%
}
\vspace{-10pt}
\end{table}

\section{Conclusion}
\label{paper:conclusion}
In this paper, we analyze the access pattern of kernel value caching in SVM training. We find that the kernel values tend to have similar access frequency across different stages of the training process, while the proportion of small reuse intervals is relatively large in the late stage. Thus, we propose HCST, an adaptive caching strategy for kernel values caching in the SVM training, together with a particularly optimization for multi-output learning tasks. The key ideas of HCST are as follows. (i) EFU is a new frequency-based strategy enhancing LFU, which stores kernel values that have higher used frequencies (ii) HCST uses a dynamic selection scheme to switch the caching strategy between EFU and LRU in the training process. According to our empirically studies, HCST mostly has a better hit ratio than the widely used caching strategy LRU, and the other existing strategies such as LFU and LAT in the SVM training. ThunderSVM can get a significant speedup by applying the HCST strategy with small cost on memory.

\section{Acknowledgements}
This work is supported by a MoE AcRF Tier 1 grant (T1 251RES1824) and a MOE Tier 2 grant (MOE2017-T2-1-122) in Singapore.

\ifCLASSOPTIONcaptionsoff
  \newpage
\fi

\bibliographystyle{IEEEtran}
\bibliography{IEEEfull}


\vspace{-20pt}
\begin{IEEEbiography}[{\includegraphics[width=1in,height=1.25in, clip,keepaspectratio]{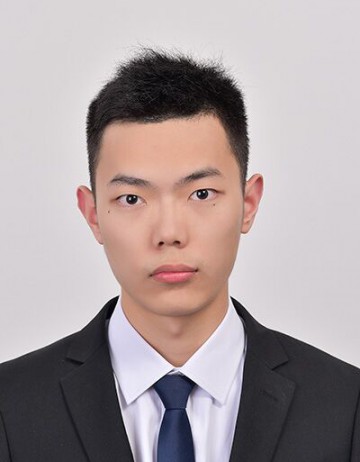}}]{Qinbin Li}
is currently a Ph.D. student with National University of Singapore.  His current research interests include machine learning, high-performance computing and privacy.
\end{IEEEbiography}
\vspace{-20pt}
\begin{IEEEbiography}
[{\includegraphics[width=1in,height=1.25in, clip,keepaspectratio]{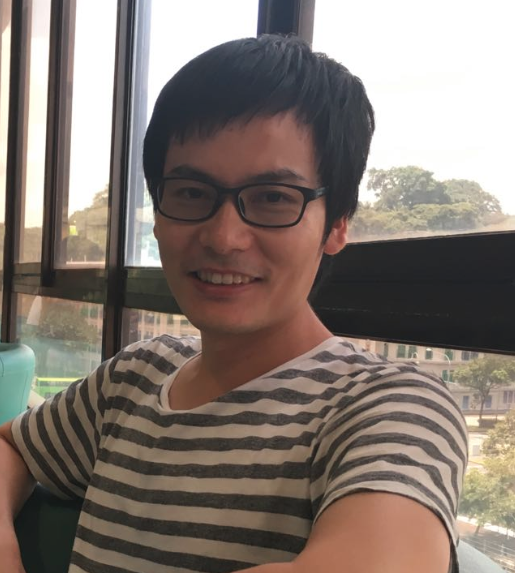}}]{Zeyi Wen}
is a Lecturer of Computer Science in The University of Western Australia. He received his PhD degree in Computer Science from The University of Melbourne in 2015. Zeyi’s areas of research include machine learning systems, automatic machine learning, high-performance computing and data mining.
\end{IEEEbiography}
\vspace{-20pt}
\begin{IEEEbiography}
[{\includegraphics[width=1in,height=1.25in, clip,keepaspectratio]{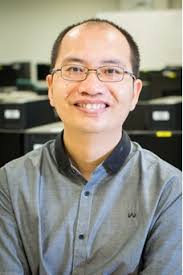}}]{Bingsheng He}
is an Associate Professor in School of Computing of National University of Singapore. He received the bachelor degree in computer science from Shanghai Jiao Tong University (1999-2003), and the PhD degree in computer science in Hong Kong University of Science and Technology (2003-2008). His research interests are high performance computing, distributed and parallel systems, and database systems.
\end{IEEEbiography}
\vfill


\end{document}